\documentclass[11pt]{article}

\usepackage[preprint]{acl}

\usepackage{times}
\usepackage{latexsym}
\usepackage{hyperref}
\usepackage{url}
\usepackage{graphicx}
\usepackage{booktabs}
\usepackage{multirow}
\usepackage{threeparttable}
\usepackage[table]{xcolor}
\usepackage{algorithm}
\usepackage{algpseudocode}
\usepackage{array}
\usepackage{tabularx}
\usepackage{natbib}
\usepackage{listings}
\usepackage{amssymb} 
\usepackage[table]{xcolor}
\usepackage{adjustbox}
\usepackage{wrapfig}

\usepackage[T1]{fontenc}

\usepackage[utf8]{inputenc}

\usepackage{microtype}

\usepackage{inconsolata}

\usepackage{graphicx}

\title{Omni-Workspace: Benchmarking Omni-Modal AI Assistants on Multimodal Workspace Tasks}

\author{
  \textbf{Yuhao Wang\textsuperscript{1}},
  \textbf{Yucheng Zhao\textsuperscript{2}},
  \textbf{Yiran Zhong\textsuperscript{2,*}},
  \textbf{Feida Zhu\textsuperscript{2}},
  \textbf{Hongcheng Liu\textsuperscript{1}},
  \\
  \textbf{Qichao Wang\textsuperscript{3}},
  \textbf{Ming Ma\textsuperscript{4}},
  \textbf{Yu Wang\textsuperscript{1,*}}
  \\[0.8em]
  \textsuperscript{1}Shanghai Jiao Tong University \quad
  \textsuperscript{2}Tongyi Lab, Alibaba Group \\
  \textsuperscript{3}Nanyang Technological University \\
  \textsuperscript{4}University of Chinese Academy of Sciences \\
  \textsuperscript{*}Corresponding author
}

\begin{document}
\maketitle
\begin{abstract}
Omni-modal agents are expected to move beyond multimodal understanding toward completing complex tasks in realistic multimodal  environments. In these settings, information is distributed across heterogeneous files and modalities, with cross-file and cross-modal dependencies requiring agents to continually coordinate perception, reasoning, and tool use as newly gathered evidence informs subsequent actions. However, existing omni-modal benchmarks primarily focus on multimodal perception and understanding, leaving complex task completion through sustained interaction with multimodal workspaces underexplored. We introduce \textbf{Omni-Workspace}, a benchmark comprising 380 deep-research-style tasks across four multimodal workspace types and two difficulty levels. These tasks require agents to inspect and process workspace files, retrieve external information when needed, integrate evidence across files and modalities, and produce artifacts satisfying explicit requirements. Evaluating 18 representative models reveals limitations in long-horizon multimodal workspace execution, with the best model scoring only 65.93 on average. Further analysis identifies multimodal context inflation and task-state drift as recurring challenges. Motivated by these findings, we further propose \textbf{OmniFocus}, a lightweight approach combining multimodal working-memory management with explicit task-state tracking to improve performance and efficiency. Omni-Workspace advances the evaluation and development of omni-modal agents toward autonomous completion of complex tasks in multimodal workspaces.
\end{abstract}

\section{Introduction}
\label{sec:introduction}

\begin{figure*}[t]
    \centering
    \includegraphics[
        width=\textwidth,
        trim=0 26 0 0,
        clip
    ]{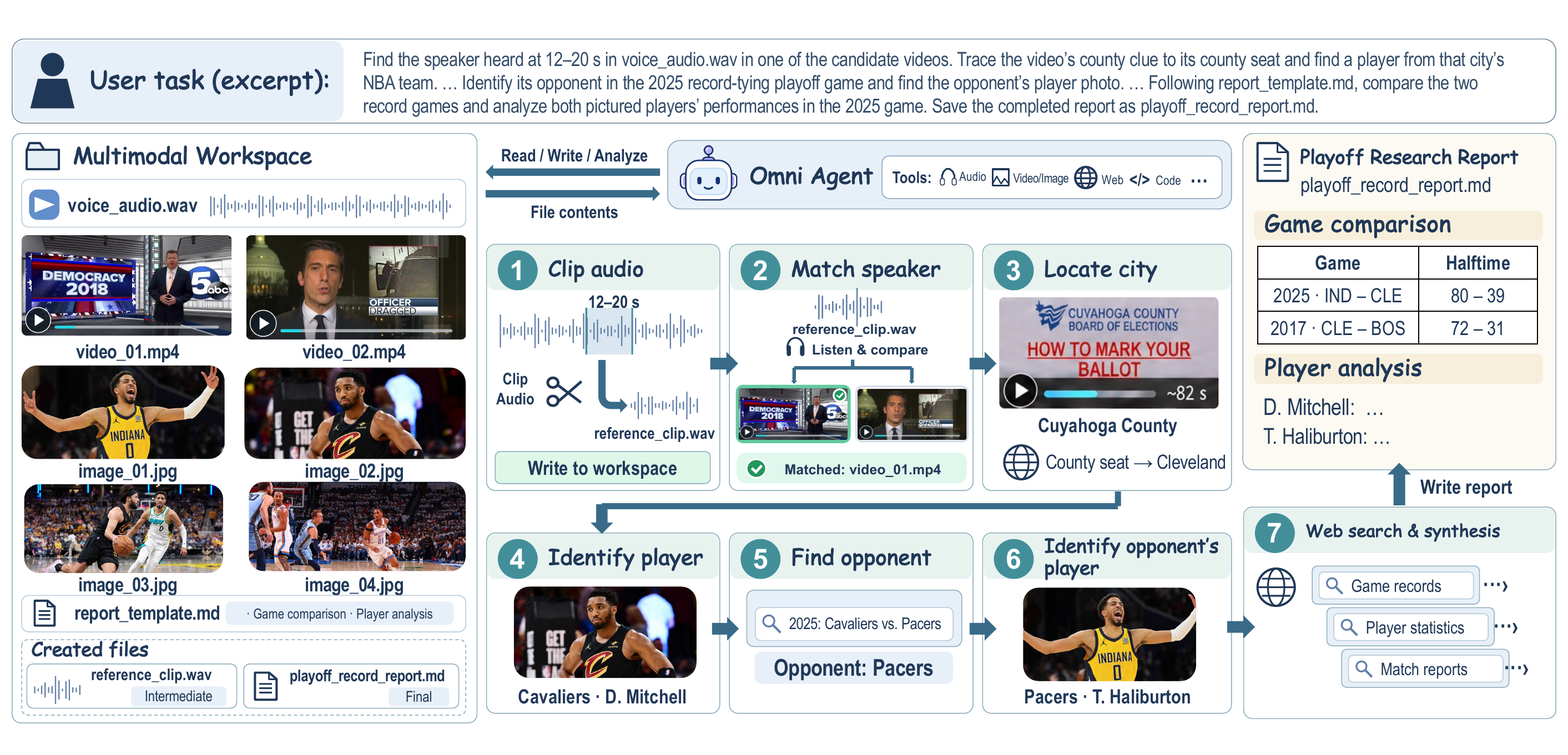}
    \caption{
    \textbf{Overview of Omni-Workspace.}
    An omni-modal agent continually coordinates perception, reasoning, and tool use to gather, process, and integrate information across workspace files and modalities, producing artifacts that meet task requirements.
    The example combines multimodal analysis, web search, and intermediate file creation and reuse to produce a structured research report.
}
    \label{fig:omni_workspace_overview}
\end{figure*}

Recent advances in omni-modal large language models (Omni-LLMs) enable unified processing of textual, visual, and auditory information~\citep{vita152025,baichuanomni152025,qwen25omni2025,qwen3omni2025,qwen35omni2026,gemini31pro2026}, providing a foundation for agents that perceive and reason over multimodal content~\citep{agentomni2025,omnigaia2026,nativeomniagent2026}. Real-world applications, however, require these agents to go beyond multimodal understanding and complete complex tasks in multimodal environments. In applications such as Google Gemini Notebook, users bring together documents and media in a workspace for analysis and synthesis~\citep{gemininotebook2026}. For example, producing structured meeting minutes may require an agent to locate key discussions in meeting recordings and connect them with presentation slides and chat logs. An ambiguous reference in a recording may prompt the agent to consult the slides or subsequent messages before drawing a conclusion. Such tasks require agents to continually coordinate perception, reasoning, and tool use within multimodal workspaces to gather, process, and integrate information across files and modalities and produce artifacts that satisfy user requirements.

Yet existing omni-modal benchmarks primarily focus on multimodal perception and understanding~\citep{omnieval2025,omnivideobench2025,lvomnibench2026}. OmniGAIA extends evaluation to multimodal reasoning and multi-step tool use~\citep{omnigaia2026}, while Omni-DeepSearch combines multimodal clue extraction with cross-modal search to produce concise answers~\citep{omnideepsearch2026}. 
Both provide multimodal inputs directly in the initial context, rather than requiring agents to actively discover and process multiple multimodal files in the workspace to extract and integrate information across files and modalities.
Workspace-Bench evaluates workspace tasks with cross-file dependencies~\citep{workspacebench2026}, but offers limited coverage of audio and video. It remains underexplored whether omni-modal agents can complete complex tasks through sustained interaction with persistent multimodal workspaces, integrating evidence across files and modalities to produce artifacts that satisfy explicit requirements.

To address this gap, we introduce \textbf{Omni-Workspace}, a benchmark of 380 deep-research-style tasks for evaluating omni-modal agents on cross-file and cross-modal information processing in multimodal workspaces. It spans two difficulty levels (L1 and L2), four workspace types, and 10 topics. Each task pairs a user instruction with a workspace containing files such as audio, video, images, and documents. As shown in Figure~\ref{fig:omni_workspace_overview}, agents must coordinate perception, reasoning, and tool use to autonomously discover, inspect, and process relevant files and search the open web for supporting information. They must continually integrate evidence across files and modalities to guide subsequent actions and produce artifacts meeting task requirements. Construction involves collecting and annotating publicly available media, assembling workspaces through cross-file and cross-modal links, and designing tasks with multi-stage verification for validity and solvability.

Evaluating omni-modal agents on Omni-Workspace reveals two challenges. First, raw multimodal observations can accumulate in context across turns, leading to \emph{multimodal context inflation} and increased processing overhead. Second, agents may lose track of task requirements, collected evidence, unresolved information needs, and execution progress as these become scattered across the growing interaction history, resulting in \emph{task-state drift}. These challenges are accompanied by repeated file inspection, overlooked information gaps, unnecessary searches, and premature or delayed termination. Motivated by these observations, we propose \textbf{OmniFocus}, a lightweight extension to the agent harness that combines multimodal working-memory management with explicit task-state tracking to mitigate both challenges and improve task performance and execution efficiency.

We evaluate 18 representative models under a unified agent harness, including native omni-modal models and vision-language models (VLMs) equipped with audio tools. Omni-Workspace remains challenging: the best-performing model achieves an average score of 65.93, while most models score below 60. OmniFocus also substantially reduces execution overhead on representative backbone models while improving task performance. These results suggest that managing multimodal context and explicitly tracking task state can help agents complete complex workspace tasks more effectively and efficiently. Our main contributions are threefold:
\begin{itemize}

    \item We introduce \textbf{Omni-Workspace}, a benchmark of 380 deep-research-style tasks for evaluating omni-modal agents in multimodal workspaces. These tasks require agents to coordinate perception, reasoning, and tool use across turns to gather, process, and integrate information across files and modalities and produce artifacts meeting task requirements.

    \item We systematically evaluate 18 representative models, revealing agents' limitations in completing complex multimodal workspace tasks and identifying multimodal context inflation and task-state drift as recurring challenges.


    \item We analyze execution failures and identify recurring challenges in context management and task-state maintenance. These findings inform OmniFocus, a lightweight harness enhancement whose evaluation illustrates practical opportunities to improve task completion and execution efficiency.
\end{itemize}

\section{Related Work}
\label{sec:related_work}

Omni-modal benchmarks have expanded from media understanding to tool-assisted problem solving. OmniEval, OmniVideoBench, and LVOmniBench assess multimodal comprehension and reasoning~\citep{omnieval2025,omnivideobench2025,lvomnibench2026}, while OmniGAIA and Omni-DeepSearch incorporate multi-step tool use and external evidence retrieval~\citep{omnigaia2026,omnideepsearch2026}. Despite differing reasoning and tool requirements, these benchmarks largely organize tasks around multimodal inputs supplied with individual questions. However, agents carrying out complex tasks in practical environments need to work with information distributed across multiple files. They must decide what to inspect and process as the task unfolds, integrating findings across files and modalities to fulfill the user's request. Workspace-Bench takes a step toward evaluating such task execution through contextual reasoning over interdependent workspace files~\citep{workspacebench2026}, but offers limited coverage of audio and video. Omni-Workspace examines this setting for omni-modal agents, requiring them to acquire, process, and integrate information across multimodal workspace files to produce task-specified artifacts.

\section{Omni-Workspace}
\label{sec:benchmark}

\begin{figure*}[t]
    \centering
    \includegraphics[
        width=\textwidth,
        trim=0 0 0 0,
        clip
    ]{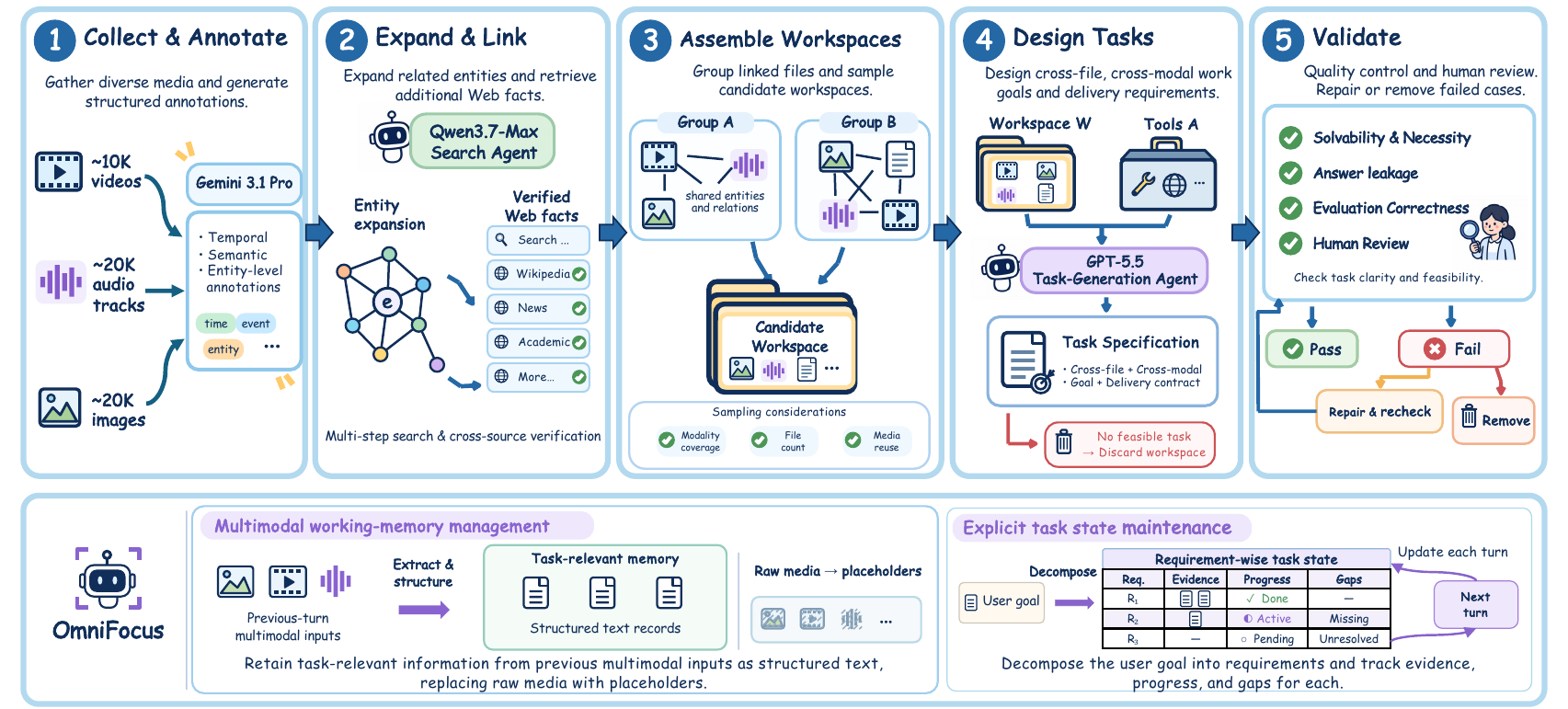}
    \caption{
    \textbf{Omni-Workspace construction and OmniFocus.}
    \textbf{Top:} The five-stage dataset construction pipeline.
    \textbf{Bottom:} OmniFocus combines multimodal working-memory management with explicit task-state maintenance.
}\label{fig:omni_workspace_construction}
\end{figure*}

\subsection{Task Formulation}
\label{sec:task_formulation}

Each Omni-Workspace task pairs a natural-language user goal with a multimodal workspace and requires one or more target files. We formulate a task as
\begin{equation}
    \mathcal{T} = (G, W_0, \mathcal{A}, R),
\end{equation}
where $G$ specifies the task objective and delivery requirements, including formats and required content; $W_0$ is the initial workspace containing heterogeneous multimodal files; $\mathcal{A}$ is the task-specific tool and action space; and $R$ is a task-specific evaluation rubric hidden from the agent. The agent receives $G$ and accesses workspace contents through tool calls. At step $t$, executing an action $a_t \in \mathcal{A}$ returns an observation for subsequent decisions and may modify the workspace:
\begin{equation}
    W_{t+1} = F_W(W_t, a_t),
\end{equation}
where $W_t$ denotes the workspace file state and $F_W$ captures the effects of tool execution. Tasks are designed to require information integration across files and modalities, supplemented by information from the open Web, to produce target files. At termination, the target files are collected from the specified paths and evaluated against $R$.

\subsection{Data Collection and Task Construction}
\label{sec:data_task_construction}

Figure~\ref{fig:omni_workspace_construction} summarizes the construction pipeline.

\paragraph{Multimodal source pool.}
We build an initial source pool comprising approximately 10K videos from public video platforms and FineVideo~\citep{farre2024finevideo}, 20K audio tracks extracted from FineVideo videos, and 20K images from LiveVQA~\citep{fu2025livevqa} and COCO 2017~\citep{lin2014coco}. Gemini 3.1 Pro~\citep{gemini31pro2026} generates modality-specific structured annotations that capture temporal, semantic, and entity-level information.
Using the media annotations as search cues, a Search Agent based on Qwen3.7-Max~\citep{qwen37max2026} identifies related people, organizations, locations, works, and events through Web search and browsing. It also collects verifiable facts about these entities to support subsequent workspace construction and task design. The Search Agent is instructed to prioritize information that requires multiple search steps to obtain and to avoid facts that are common knowledge or directly available in a single search-result snippet.

\paragraph{Entity-linked workspace construction.}
We normalize entity mentions across the annotated media and retrieved web information, and establish cross-file and cross-modal links through shared entities and semantic relations. Related files are organized into groups based on these links, and groups with compatible entities and relations are combined into candidate workspaces. Sampling considers modality coverage, workspace size, and media reuse, while limiting repeated sampling of frequently occurring entities or files.

\paragraph{Workspace-grounded task design.}
We use GPT-5.5~\citep{openai2026gpt55} as the Task-Generation Agent. For each candidate workspace, the agent receives media annotations, cross-file relationships, retrieved Web information, and specifications of the tool subset available to the evaluated agent. It uses these relationships to design a task requiring information integration across files and modalities, supplemented by information retrieved from the open Web. The agent generates a user instruction specifying the task objective and delivery requirements for target files. It also generates gold result files and a task-specific evaluation rubric for subsequent validation. Candidate workspaces are discarded when the generator cannot formulate a task meeting these requirements with the assigned tools. Further details are provided in Appendix~\ref{app:construction-details}.

\paragraph{Quality control.}
Candidate tasks undergo three types of LLM-based checks.
\emph{Solvability and necessity} checks assess whether the required information can be obtained from the workspace and open web using the provided tools, and whether the specified files, modalities, and tool contribute substantively to task completion.
\emph{Answer leakage} checks inspect user instructions, filenames, metadata, and templates for unintended answer disclosure.
\emph{Evaluation correctness} checks assess whether the gold result files are correct and satisfy the task instruction. They also examine whether the rubric covers the required content and format, with clear scoring criteria consistent with these files. Three annotators with graduate-level research experience subsequently review task clarity, feasibility, and rubric quality, including the alignment of scoring criteria with task requirements. Issues identified at either stage are addressed through revision by the task-generation agent and subsequent rechecking; tasks are discarded when reliable repair is infeasible. Together, these checks help ensure task validity and consistency among task requirements and evaluation rubrics. Detailed are provided in Appendix~\ref{app:quality-control}.

\subsection{Data Statistics}
\label{sec:data_statistics}

\begin{figure*}[t]
    \centering
    \includegraphics[width=0.94\linewidth]{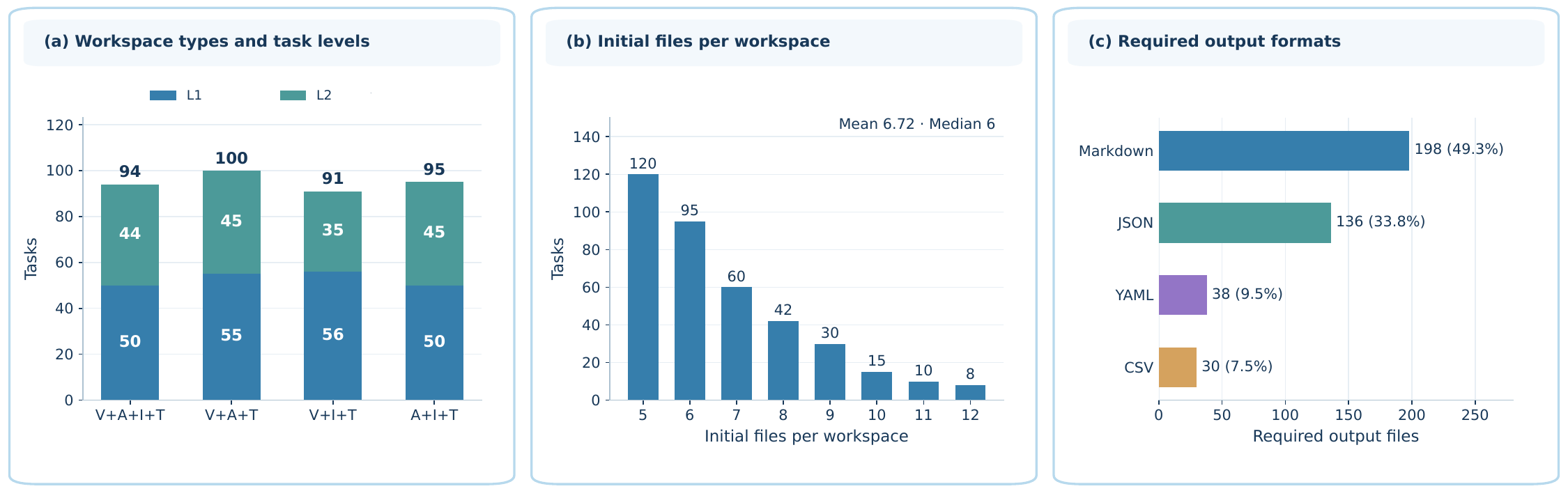}
    \caption{
        \textbf{Dataset statistics of Omni-Workspace.}
        (a) Task counts across four workspace types and two difficulty levels
        (L1 and L2).
        (b) Distribution of initial workspace file counts.
        (c) Counts and proportions of required output files by format,
        computed over 402 output files.
        V, A, I, and T denote video, audio, image, and text, respectively.
    }
    \label{fig:data_statistics}
\end{figure*}

Omni-Workspace comprises 380 tasks spanning four workspace types: video--audio--image--text (V+A+I+T), video--audio--text (V+A+T), video--image--text (V+I+T), and audio--image--text (A+I+T). Task counts are approximately balanced across these types (Figure~\ref{fig:data_statistics}(a)). We further categorize tasks by their media-processing requirements: 211 L1 tasks can be completed through direct inspection, information retrieval, and reasoning, whereas 169 L2 tasks require code execution or media-processing tools to transform or analyze media before the requested results can be obtained. The initial workspaces contain 2,552 file instances, with 5--12 files per task and an average of 6.72 (Figure~\ref{fig:data_statistics}(b)). Tasks require 402 output files in total, spanning Markdown (49.3\%), JSON (33.8\%), YAML (9.5\%), and CSV (7.5\%) (Figure~\ref{fig:data_statistics}(c)). Additional statistics and counting conventions are provided in Appendix~\ref{app:benchmark_statistics}.

\subsection{Evaluation Harness}
\label{sec:evaluation_protocol}

\paragraph{Agent harness.}

Each task runs in an isolated, persistent multimodal workspace. The agent receives the user instruction and tool schemas, and uses tools to inspect and process workspace files, retrieve web information, and produce the required output files. Tools support workspace access, multimodal perception, code execution, media processing, web search, and execution control (Appendix~\ref{app:tool-specifications}).
The harness follows an iterative reasoning--action--observation loop~\citep{yao2023react}: it executes the agent's tool calls and appends the results to the interaction context for subsequent decisions.
The agent must invoke tools to read multimodal files, and the returned media are inserted into the context as multimodal inputs for the model to inspect. Execution terminates when the agent invokes the \texttt{finish} tool or reaches the maximum number of interaction turns.

\paragraph{Evaluation protocol.}
After execution, the evaluator collects the required output files. If any required file is missing or unreadable, the entire task receives a score of zero. Otherwise, the submitted files are assessed using a task-specific rubric comprising atomic Content and Format checks. Rule-based verifiers evaluate deterministic requirements, including file types, required fields, structural constraints, and numerical results with specified tolerances. 
We use an LLM-as-a-judge approach~\citep{zheng2023judging} to assess semantic requirements. The judge evaluates each atomic check using the submitted files and the task-specific rubric, assigning a binary score: satisfied (1) or not satisfied (0).
Model settings and the agent and judge prompts are provided in Appendix~\ref{app:model-and-evaluation-settings}. Atomic check scores are aggregated within the content and format categories according to the rubric to obtain $S_{\mathrm{content}}$ and $S_{\mathrm{format}}$, both on a 0--100 scale. The overall task score combines these components using a weight $\lambda$ ($ 0 \leq \lambda \leq 1$):
\begin{equation}
\label{eqa:score}
    S_{\mathrm{overall}}
    = \lambda S_{\mathrm{content}}
    + (1-\lambda) S_{\mathrm{format}}
\end{equation}

\begin{table*}[!ht]
    \centering
    \caption{
    \textbf{Main results on Omni-Workspace.}
    Scores are task averages (0--100) computed using Equation~\ref{eqa:score} with $\lambda=0.9$.
    V, A, I, and T denote video, audio, image, and text.
    Bold indicates the highest score in each column.
}
    \label{tab:omni_workspace_main_results_no_gt_qwen37max}

    \setlength{\tabcolsep}{4.2pt}
    \renewcommand{\arraystretch}{1.08}

    \resizebox{0.91\textwidth}{!}{%
    \begin{tabular}{@{}lrrrr!{\hspace{6pt}}rrr@{}}
        \toprule
        \multirow{2}{*}{\textbf{Model}}
        & \multicolumn{4}{c!{\hspace{6pt}}}{\textbf{Workspace Type}}
        & \multicolumn{2}{c}{\textbf{Difficulty Levels}}
        & \multirow{2}{*}{\textbf{Avg.}} \\
        \cmidrule(lr){2-5}
        \cmidrule(lr){6-7}
        & \multicolumn{1}{c}{V+A+I+T}
        & \multicolumn{1}{c}{V+A+T}
        & \multicolumn{1}{c}{V+I+T}
        & \multicolumn{1}{c!{\hspace{6pt}}}{A+I+T}
        & \multicolumn{1}{c}{L1}
        & \multicolumn{1}{c}{L2}
        & \\

        \midrule
        \rowcolor{gray!12}
        \multicolumn{8}{c}{\textit{Native Omni-Modal Agents}} \\
        \midrule

        Gemini 3.5 Flash
        & \textbf{68.29} & 61.72 & 65.52 & \textbf{68.42} & \textbf{72.77} & 57.38 & \textbf{65.93} \\

        Gemini 3.1 Pro Preview
        & 65.01 & 52.43 & 58.59 & 63.93 & 65.13 & 53.36 & 59.89 \\

        Gemini 3 Flash Preview
        & 59.40 & 51.65 & 55.34 & 56.33 & 60.87 & 49.06 & 55.62 \\

        Gemini 2.5 Pro
        & 52.36 & 44.53 & 47.06 & 49.65 & 53.50 & 41.93 & 48.35 \\

        MiMo-V2.5
        & 52.40 & 42.82 & 42.11 & 48.51 & 53.02 & 38.23 & 46.44 \\

        Qwen3.5 Omni Plus
        & 42.90 & 35.27 & 39.87 & 39.76 & 44.53 & 32.95 & 39.38 \\
        
        Qwen3.5 Omni Flash
        & 26.71 & 14.92 & 28.98 & 34.01 & 30.60 & 20.20 & 25.97 \\

        MiniCPM-o-4.5
        & 14.74 & 14.96 & 11.28 & 12.96 & 15.43 & 11.14 & 13.52 \\

        Nemotron-3-Nano-Omni
        & 8.22 & 9.71 & 5.49 & 8.41 & 9.64 & 5.97 & 8.01 \\

        Qwen3-Omni-Instruct
        & 5.29 & 4.96 & 6.60 & 5.73 & 6.05 & 5.09 & 5.63 \\

        \midrule
        \rowcolor{gray!12}
        \multicolumn{8}{c}{\textit{VLM-Based Agents with ASR}} \\
        \midrule

        Qwen3.7-Plus
        & 67.08 & \textbf{62.46} & \textbf{65.88} & 63.90 & 68.70 & \textbf{59.88} & 64.78 \\

        Qwen3.6-Plus
        & 62.78 & 56.50 & 59.90 & 54.37 & 62.75 & 52.82 & 58.34 \\

        Kimi K2.6
        & 59.31 & 55.32 & 62.41 & 43.92 & 61.23 & 47.57 & 55.16 \\

        MiniMax-M3
        & 52.58 & 53.98 & 52.51 & 49.37 & 55.97 & 47.33 & 52.13 \\

        Qwen3.5-122B-A10B
        & 52.10 & 45.84 & 52.82 & 54.70 & 56.17 & 45.16 & 51.27 \\

        Qwen3.5-35B-A3B
        & 41.50 & 43.36 & 42.42 & 42.29 & 47.83 & 35.63 & 42.41 \\

        Qwen3-VL-30B-A3B-Instruct
        & 21.02 & 22.02 & 16.89 & 20.87 & 22.25 & 17.76 & 20.25 \\

        Qwen3-VL-8B-Instruct
        & 17.24 & 12.90 & 13.66 & 18.11 & 17.46 & 12.96 & 15.46 \\

        \bottomrule
    \end{tabular}%
    }
\end{table*}

\subsection{OmniFocus}
\label{sec:omnifocus}

Long-horizon tasks in Omni-Workspace require agents to inspect multimodal files and integrate information across interaction turns. Raw media returned by tools can remain in subsequent inputs after relevant information has been extracted, leading to \emph{multimodal context inflation} and increased processing overhead. Meanwhile, observations and execution progress become scattered across the interaction history, making it difficult to track completed requirements, remaining gaps, and termination conditions. This \emph{task-state drift} can lead to repeated file inspection, unnecessary searches, and delayed termination. To address these challenges, we introduce \textbf{OmniFocus}, a lightweight extension to the agent harness with two complementary mechanisms that help keep subsequent reasoning and action focused on task-relevant evidence and unfinished requirements.

\paragraph{Multimodal working-memory management.}
OmniFocus helps the agent focus on task-relevant evidence by retaining key observations from inspected media as textual records for subsequent reasoning. When tools return images, audio, or video, the harness prompts the agent to record key observations. The agent generates this record together with its next action in the same response. After this response, the harness replaces the inspected media with placeholders in subsequent contexts. This replacement affects only the interaction context; source files remain available in the workspace and can be revisited when additional details are needed.

\paragraph{Explicit task-state maintenance.}
OmniFocus maintains a structured task state to track which requirements have been satisfied and what remains to be done. The agent initializes this state from the user goal. For each requirement, it records the completion condition, acquired information, processing progress, and remaining gaps or conflicts. The agent updates the state based on new observations and execution results, and the harness includes the latest state in subsequent model inputs. This requirement-wise summary helps the agent focus on unfinished requirements and decide when to gather additional information, generate and validate output files, or terminate. Together, these mechanisms aim to improve both task performance and execution efficiency. Tool interfaces and the execution workflow are detailed in Appendix~\ref{app:omnifocus}.

\section{Experiments}
\label{sec:experiments}

\begin{figure*}[!t]
    \centering
    \includegraphics[width=0.99\linewidth]{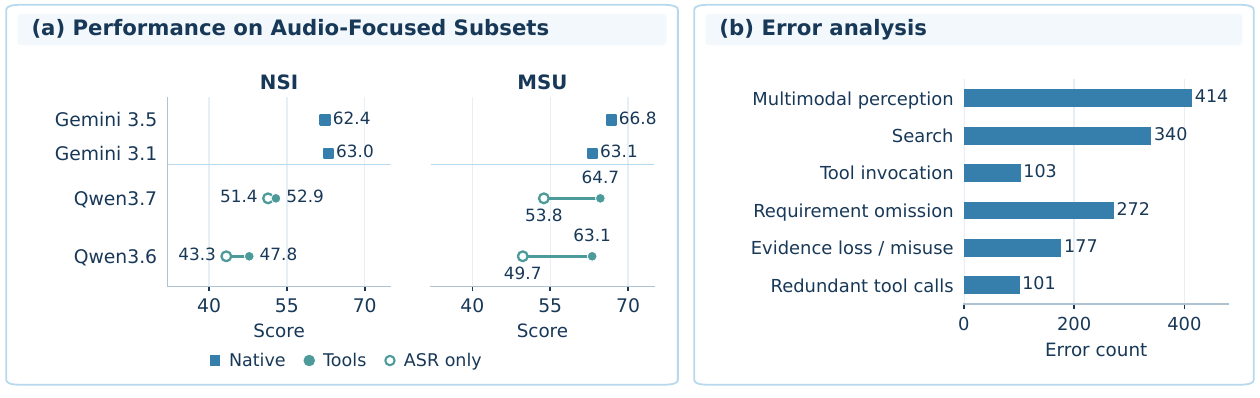}
    \caption{\textbf{Performance on audio-focused subsets and error analysis.}
(a) Scores on NSI and MSU subsets. Squares denote native omni-modal models; filled and open circles denote audio tool-equipped and ASR-only configurations, respectively.
(b) Error counts by category.}
    \label{fig:analysis-overview}
\end{figure*}

\subsection{Experimental Settings}
\label{sec:experimental_settings}

We evaluate native omni-modal agents and VLM-based agents equipped with automatic speech recognition (ASR) on all 380 tasks in Omni-Workspace. All agents use the evaluation harness described in Section~\ref{sec:evaluation_protocol}, with the same task instructions and a maximum of 50 agent steps per task. For each task, the available tools are shared across models except for audio access: native omni-modal agents use an audio-reading tool, while VLM-based agents use an ASR tool that returns speech transcriptions (Appendix~\ref{app:tool-specifications}). Deterministic checks are scored programmatically, while Qwen3.7-Max evaluates semantic checks. We compute the overall task score using Equation~\ref{eqa:score} with $\lambda=0.9$ and report mean task scores on a 0--100 scale. Model access and inference settings are detailed in Appendix~\ref{app:model-and-evaluation-settings}.

\subsection{Main Results}
\label{sec:main_results}

\paragraph{Overall performance.}
Table~\ref{tab:omni_workspace_main_results_no_gt_qwen37max} reports the main results on Omni-Workspace. Gemini 3.5 Flash achieves the highest overall score of 65.93, closely followed by Qwen3.7-Plus at 64.78. Gemini 3.1 Pro Preview and Qwen3.6-Plus rank next, scoring 59.89 and 58.34, respectively. Only the top two models score above 60, indicating that Omni-Workspace remains challenging for current agents.

\paragraph{Native omni-modal agents and VLM+ASR agents.}

The strongest agents from the two categories achieve comparable overall performance, with different strengths across task subsets. Qwen3.7-Plus leads on V+A+T, V+I+T, and L2, whereas Gemini 3.5 Flash leads on the other two workspace types and L1. Among native omni-modal agents, the strongest performance is concentrated in the Gemini family; other models, particularly smaller open-weight models, score substantially lower. 
Although these models support native multimodal perception, they struggle to coordinate perception and tool invocation over multiple steps to produce the required deliverables. 

\paragraph{Performance across difficulty levels.}

All evaluated models score lower on L2 than on L1. This gap is also evident among the strongest agents: Gemini 3.5 Flash scores 72.77 on L1 and 57.38 on L2, while Qwen3.7-Plus scores 68.70 and 59.88, respectively. L2 tasks require agents to transform or analyze media using code and media-processing tools and integrate the resulting information into subsequent reasoning. The consistently lower L2 scores indicate that these processing-dependent tasks remain more challenging for current agents.

\paragraph{Performance across workspace types.}
No single workspace type consistently yields the lowest scores across models. A notable contrast emerges on A+I+T, where native omni-modal agents generally maintain relatively strong performance, while several VLM+ASR agents obtain their lowest workspace scores. Gemini 3.5 Flash achieves its highest workspace score of 68.42 on A+I+T, whereas Qwen3.6-Plus, Kimi K2.6, and MiniMax-M3 obtain their lowest scores on this subset, at 54.37, 43.92, and 49.37, respectively. This pattern motivates a closer comparison of native omni-modal and VLM-based agents on audio-focused tasks in Section~\ref{sec:audio_analysis}.

\section{Analysis}
\label{sec:analysis}

\subsection{Native Omni-Modal Agents on Audio-Focused Tasks}
\label{sec:audio_analysis}

\begin{table*}[t]
\centering
\caption{Performance and efficiency of OmniFocus and its ablations.
Best values within each model are bold.
Costs are estimated using uniform standard-tier token rates, without audio
surcharges, long-context pricing tiers, or cache discounts.}
\label{tab:omnifocus-results}
\small
\setlength{\tabcolsep}{3pt}
\renewcommand{\arraystretch}{1.0}
\begin{adjustbox}{max width=0.99\linewidth}
\begin{tabular}{@{}llrrrrrrr@{}}
\toprule
Model & Setting & Score $\uparrow$ & Tool Calls $\downarrow$
& Turns $\downarrow$ & Input Tokens (k) $\downarrow$
& Output Tokens (k) $\downarrow$ & Cost (\$) $\downarrow$ & Time (s) $\downarrow$ \\
\midrule
\multirow{4}{*}{Gemini 3 Flash Preview}
& OmniFocus & \textbf{60.73} & 25.62 & 9.35 & \textbf{249.20} & 7.38 & \textbf{0.1467} & \textbf{218.38} \\
& \quad w/o task state & 57.10 & 32.17 & 11.46 & 291.04 & 5.59 & 0.1623 & 336.39 \\
& \quad w/o mem management & 60.05 & \textbf{25.43} & \textbf{9.14} & 426.53 & 6.33 & 0.2323 & 550.94 \\
& \quad w/o both & 55.62 & 27.63 & 10.68 & 330.57 & \textbf{3.24} & 0.1750 & 301.19 \\
\midrule
\multirow{4}{*}{Gemini 3.1 Pro Preview}
& OmniFocus & \textbf{64.95} & \textbf{27.62} & \textbf{10.40} & \textbf{271.07} & 16.14 & \textbf{0.7358} & 564.08 \\
& \quad w/o task state & 63.23 & 29.51 & 11.69 & 299.87 & \textbf{12.92} & 0.7548 & \textbf{550.88} \\
& \quad w/o mem management & 64.20 & 30.38 & 11.65 & 580.11 & 18.53 & 1.3826 & 821.65 \\
& \quad w/o both & 59.89 & 31.32 & 15.00 & 615.81 & 12.98 & 1.3874 & 645.25 \\
\bottomrule
\end{tabular}
\end{adjustbox}
\end{table*}

We compare representative native omni-modal agents and VLM-based agents on two audio-focused subsets: Non-Speech Information (NSI), which requires interpreting acoustic cues beyond speech content, such as timbre, environmental sounds, and music; and Multimodal Speaker Understanding (MSU), which requires associating spoken content with the correct speakers across segments.As shown in Figure~\ref{fig:analysis-overview}, the evaluated native omni-modal agents outperform the compared VLM-based agents on NSI, while VLM-based agents equipped with ASR and additional audio-analysis capabilities achieve comparable performance on MSU. Restricting Qwen3.6-Plus to ASR-only by removing both speaker-analysis tools and Python-based audio analysis reduces its score by 13.38 points on MSU and 4.43 points on NSI. Together, these results highlight the strengths of the evaluated native omni-modal agents on tasks requiring non-speech acoustic evidence, while showing that additional audio-analysis capabilities benefit VLM-based agents particularly on speaker-focused tasks. See Appendix~\ref{app:audio_analysis_details} for more details.

\subsection{Error Analysis}
\label{sec:error-analysis}

To understand how agents fail in long-horizon multimodal tasks, we analyze two types of errors: \emph{local execution errors} and \emph{task-state drift errors}. Local execution errors concern individual operations, including perception, search, and tool invocation. Task-state drift errors concern failures to maintain requirements, evidence, and execution progress across reasoning steps, manifested as requirement omission, evidence loss or misuse, and redundant tool calls. As shown in Figure~\ref{fig:analysis-overview}(b), multimodal perception and search errors dominate local execution failures, while requirement omission is the most common form of task-state drift. Evidence loss or misuse and redundant tool calls reveal failures that extend across steps: agents may misapply acquired evidence or continue using tools without substantive progress. These findings highlight the need to track requirements, retain and correctly use evidence, and update actions based on execution outcomes, providing motivation for the design of OmniFocus. More details are provided in Appendix~\ref{app:error_analysis_details}.

\subsection{OmniFocus: Effectiveness and Efficiency}
\label{sec:omnifocus-analysis}

We evaluate OmniFocus on Gemini 3 Flash and Gemini 3.1 Pro against three ablations: without task-state maintenance, without memory management, and without both components (the baseline). As shown in Table~\ref{tab:omnifocus-results}, full OmniFocus achieves the highest reported score and the lowest input-token use and estimated cost for both models. Relative to the baseline, scores increase by 5.11 and 5.06 points, input tokens decrease by 24.6\% and 56.0\%, and estimated costs decrease by 16.2\% and 47.0\%, respectively. The two ablations reveal distinct contributions. Removing task-state maintenance lowers scores by 3.63 and 1.72 points and increases tool calls from 25.62 to 32.17 for Gemini 3 Flash and from 27.62 to 29.51 for Gemini 3.1 Pro. This pattern is consistent with explicit task state helping agents organize actions around task requirements. Removing memory management has a smaller effect on score (0.68 and 0.75 points), but raises input-token use by 71.2\% and 114.0\% relative to full OmniFocus. It also raises estimated cost by 58.4\% and 87.9\%, respectively, suggesting that compacting multimodal observations primarily controls context and expense. Detailed comparisons are provided in Appendix~\ref{app:omnifocus_analysis_details}.

\section{Conclusion}
\label{sec:conclusion}

We introduced \textbf{Omni-Workspace}, a benchmark of 380 deep-research-style tasks requiring omni-modal agents to continually coordinate perception, reasoning, and tool use in multimodal workspaces. Agents must gather, process, and integrate information across files and modalities, retrieve supporting web information, and produce artifacts meeting explicit requirements. Evaluation of 18 representative models reveals substantial room for improvement, with a best average score of 65.93. Our analysis identifies multimodal context inflation and task-state drift as recurring challenges, motivating {OmniFocus}, which combines multimodal working-memory management with explicit task-state tracking. OmniFocus improves scores while substantially reducing input-token use and estimated cost on both evaluated backbones. We hope Omni-Workspace supports further evaluation and development of omni-modal agents capable of completing complex tasks in multimodal workspaces.

\section*{Limitations}

There are several limitations in this work. First, although the benchmark requires sustained interaction with multimodal workspaces, the current workspaces remain moderate in scale, containing 5--12 initial files per task. As a result, they do not yet capture the full range of challenges that may arise in much larger and noisier real-world workspaces containing hundreds or thousands of files. Second, Omni-Workspace primarily focuses on deep-research-style tasks that require agents to gather, process, verify, and synthesize information across multimodal files and the Web before producing structured artifacts. In real-world applications, omni-modal agents may also support a broader range of workflows beyond information-centric tasks, such as video editing, multimedia content creation, game development, and other forms of multimodal artifact creation and manipulation. Given the substantially greater complexity of these workflows and the challenges of constructing and reliably validating them at scale, Omni-Workspace does not currently cover these task types.

\section*{Ethical Considerations}

Omni-Workspace is constructed from publicly available datasets and media sources, including FineVideo, LiveVQA, and COCO 2017, together with supporting information retrieved from the public Web. We will release the benchmark and associated resources in accordance with the applicable licenses and terms of use, preserving required attribution and respecting restrictions on reuse and redistribution. Original media will be redistributed only where permitted by the corresponding licenses or permissions; otherwise, we will provide references to the original sources. The annotators were informed of the research purpose and how their feedback would be used. No personal identifiers of the annotators will be included in the public release. Omni-Workspace is intended to support research on multimodal workspace agents. We used large language models to assist with editing and polishing the manuscript. The authors remain responsible for the final content.




\bibliography{custom}

\appendix

\appendix

\section{Comparison with Existing Benchmarks}
\label{app:benchmark_comparison}

Table~\ref{tab:multimodal_benchmark_comparison} compares representative benchmarks for omni-modal understanding, multimodal agents, and workspace task execution along four dimensions: modality coverage, task dependencies, agentic operations, and output forms.

Existing omni-modal understanding benchmarks primarily evaluate perception and reasoning over task-provided multimodal inputs, while multimodal agent benchmarks further incorporate multi-step reasoning, external information retrieval, and tool use. Workspace-Bench extends evaluation to persistent file environments, emphasizing cross-file dependencies and workspace-level task completion, but provides limited coverage of audio, video, and cross-modal media processing.

Omni-Workspace focuses on dependency-driven execution across heterogeneous multimodal files. Its tasks require agents to integrate evidence across files and modalities, perform media-processing operations when needed, and reuse intermediate artifacts in subsequent steps. By combining these requirements with Web retrieval and structured artifact generation, Omni-Workspace evaluates sustained coordination of perception, reasoning, and tool use in multimodal workspaces.

\begin{table*}[t]
\centering
\caption{
Comparison of representative benchmarks by modality coverage,
task dependencies, agentic operations, and output form.
}
\label{tab:multimodal_benchmark_comparison}

\small
\setlength{\tabcolsep}{4.5pt}
\renewcommand{\arraystretch}{1.12}

\resizebox{\textwidth}{!}{%
\begin{tabular}{l c c c c c c c c l}
\toprule

\multirow{3}{*}{\textbf{Benchmark}}
& \multicolumn{3}{c}{\textbf{Modalities}}
& \multicolumn{2}{c}{\textbf{Task Dependencies}}
& \multicolumn{3}{c}{\textbf{Agentic Operations}}
& \multirow{3}{*}{\textbf{Output Form}} \\

\cmidrule(lr){2-4}
\cmidrule(lr){5-6}
\cmidrule(lr){7-9}

& \multirow{2}{*}{\textbf{Aud.}}
& \multirow{2}{*}{\textbf{Vid.}}
& \multirow{2}{*}{\textbf{Img.}}
& \multirow{2}{*}{\shortstack{\textbf{Cross-file}\\\textbf{Dependency}}}
& \multirow{2}{*}{\shortstack{\textbf{Intermediate}\\\textbf{Artifact Reuse}}}
& \multirow{2}{*}{\shortstack{\textbf{Web}\\\textbf{Retrieval}}}
& \multirow{2}{*}{\shortstack{\textbf{Media}\\\textbf{Processing}}}
& \multirow{2}{*}{\shortstack{\textbf{Workspace}\\\textbf{File I/O}}}
& \\

& & & & & & & & & \\[-3pt]

\midrule

OmniEval
& $\checkmark$
& $\checkmark$
& $\times$
& $\times$
& $\times$
& $\times$
& $\times$
& $\times$
& Short-form Answer \\

OmniVideoBench
& $\checkmark$
& $\checkmark$
& $\times$
& $\times$
& $\times$
& $\times$
& $\times$
& $\times$
& Short-form Answer \\

LVOmniBench
& $\checkmark$
& $\checkmark$
& $\times$
& $\times$
& $\times$
& $\times$
& $\times$
& $\times$
& Short-form Answer \\

\midrule

OmniGAIA
& $\checkmark$
& $\checkmark$
& $\checkmark$
& $\times$
& $\times$
& $\checkmark$
& $\times$
& $\checkmark$
& Short-form Answer \\

Omni-DeepSearch
& $\checkmark$
& $\checkmark$
& $\checkmark$
& $\times$
& $\times$
& $\checkmark$
& $\times$
& $\times$
& Short-form Answer \\

MM-BrowseComp
& $\times$
& $\checkmark$
& $\checkmark$
& $\times$
& $\times$
& $\checkmark$
& $\times$
& $\times$
& Short-form Answer \\

VideoDR
& $\times$
& $\checkmark$
& $\times$
& $\times$
& $\times$
& $\checkmark$
& $\times$
& $\times$
& Short-form Answer \\

\midrule

Workspace-Bench
& $\times$
& $\times$
& $\checkmark$
& $\checkmark$
& $\checkmark$
& $\times$
& $\times$
& $\checkmark$
& Target File \\

\midrule

\textbf{Omni-Workspace}
& $\checkmark$
& $\checkmark$
& $\checkmark$
& $\checkmark$
& $\checkmark$
& $\checkmark$
& $\checkmark$
& $\checkmark$
& \textbf{Target File} \\

\bottomrule
\end{tabular}%
}

\vspace{2pt}

\begin{minipage}{0.99\textwidth}
\footnotesize
\textit{Definitions.}
\textbf{Cross-file Dependency}: evidence from one file guides the interpretation or selection of another.
\textbf{Intermediate Artifact Reuse}: generated files are reused in subsequent operations.
\textbf{Web Retrieval}: retrieving external information from the Web.
\textbf{Media Processing}: analyzing or transforming media files beyond direct perception.
\textbf{Workspace File I/O}: tool-based access to or modification of task-related files.
\textbf{Target File}: a task-specified file deliverable.
Checkmarks indicate benchmark task requirements.
Text is omitted as it is supported by all benchmarks.
\end{minipage}

\end{table*}

\section{Dataset Construction and Quality Control}
\label{app:dataset_construction}

\subsection{Construction Details}
\label{app:construction-details}

\paragraph{Media annotation.}
We use modality-specific annotation procedures to capture both semantic content and perceptual cues, and organize the results into structured entity records for subsequent linking. Videos are first segmented into scenes using PySceneDetect\footnote{\url{https://www.scenedetect.com/}}. Each segment receives a chronological description integrating visual events, visible text, speech, and non-speech sounds. Segment descriptions are then combined with their timestamps into a video-level narrative, from which entities are extracted and consolidated across segments. Audio is annotated along its timeline, capturing spoken content, speaker turns, non-speech events, and acoustic characteristics. Images receive descriptions of their scenes, objects, spatial relationships, and textual or graphical content, together with entity annotations. Across modalities, entity records include descriptions, semantic tags, and searchability information to support Web-based entity expansion and cross-file association. Available reference text assists interpretation, while descriptions are grounded in the media.

\paragraph{Workspace assembly.}
We merge entity-expansion records from video, audio, and image sources and normalize entity names into shared identifiers. Files are linked through shared entities or related entities discovered during expansion. Generic and excessively frequent entities are filtered to reduce uninformative connections. We construct connected groups of two to four files and assemble candidate workspaces from one or more groups. Sampling accounts for modality coverage, link strength, file counts, overlap between groups, and per-file reuse budgets.

\paragraph{Task specification.}
Task generation is carried out through a multi-step agent workflow. The task-generation agent receives file annotations, cross-file entity links, retrieved Web evidence, and a task-specific set of tool schemas. Across successive steps, it examines these materials, selects relevant files and evidence paths, and uses available tools to inspect metadata, check computations, or create supporting files as needed. It progressively constructs an internal reference chain specifying the input file or information, tool operation, expected observation, and downstream dependency at each step. Observations from one file guide subsequent file selection, interpretation, or Web retrieval; where processing is needed, media transformations or code execution produce intermediate results consumed by later steps. Evidence from the selected files is combined through comparison, filtering, verification, or synthesis into a concrete deliverable. Candidate workspaces are discarded if they cannot support a cross-file, cross-modal task requiring additional Web information and a file deliverable with the assigned tools. For retained candidates, the agent translates the reference chain into a user instruction specifying the objective, output paths, formats, and required content without revealing the solution steps. It also creates supporting templates or constraint files when needed, expected output contents, and a corresponding rubric. Construction annotations and the reference chain remain separate from the solver-visible workspace.

\subsection{Quality Control}
\label{app:quality-control}

\paragraph{Validation procedure.}
Candidate tasks undergo structural validation and LLM-based review. Structural checks check for missing workspace files, unresolved file references, and missing reference outputs or evaluation specifications. The LLM-based review assesses the task instruction, supporting evidence, reference solution, and rubric against the three criteria in Table~\ref{tab:dataset-quality-checks}. Together, these checks help ensure that retained tasks are well-specified, solvable, and consistently evaluated.

\paragraph{Human review.} Following the automated checks, three annotators review each candidate task for clarity and rubric quality. Clarity concerns whether the task objective and delivery requirements are sufficiently clear. Feasibility concerns whether the requested result can be obtained from the workspace and the Web using the assigned tools. Rubric quality concerns the alignment of the evaluation criteria with the task requirements and the clarity of the conditions used to determine whether each criterion is satisfied. Each annotator assigns one overall pass/fail decision to the task. A task passes human review when at least two of the three annotators assign a pass decision.

\begin{table}[t]
\centering
\small
\caption{Quality-control criteria for candidate tasks.}
\label{tab:dataset-quality-checks}
\renewcommand{\arraystretch}{1.12}
\begin{tabular}{@{}p{0.23\linewidth}p{0.72\linewidth}@{}}
\hline
\textbf{Check} & \textbf{Acceptance criteria} \\
\hline
Solvability and necessity & Required information is supported by workspace or Web evidence and obtainable with the assigned tools. Cross-file and cross-modal dependencies contribute to the requested result. For L2 tasks, processing produces information needed for completion. \\
Answer leakage & Instructions, filenames, metadata, and supporting documents do not disclose target conclusions or intermediate identities that the agent is expected to infer. \\
Evaluation correctness & The reference output is supported by evidence and satisfies the requested content and format. Rubric checks correspond to explicit requirements, accommodate equivalent correct answers, and introduce no hidden requirements or unnecessary restrictions on solution paths. \\
\hline
\end{tabular}
\end{table}

\paragraph{Repair and re-evaluation.}
Detected problems are addressed by revising the instruction, supporting files, reference output, or rubric as appropriate. Changes are propagated to dependent components so that the task remains consistent: for example, revising a requested quantity requires updating both its reference value and the associated scoring check. Repaired tasks are re-evaluated against the affected criteria. Tasks are removed when missing evidence, unsupported operations, or unresolved ambiguity prevents reliable repair.

\subsection{Benchmark Statistics and Distributions}
\label{app:benchmark_statistics}
We supplement the main-text overview with workspace and media statistics, followed by the topic coverage and instruction characteristics of all 380 tasks.

\begin{figure*}[!ht]
    \centering
    \includegraphics[width=\linewidth]{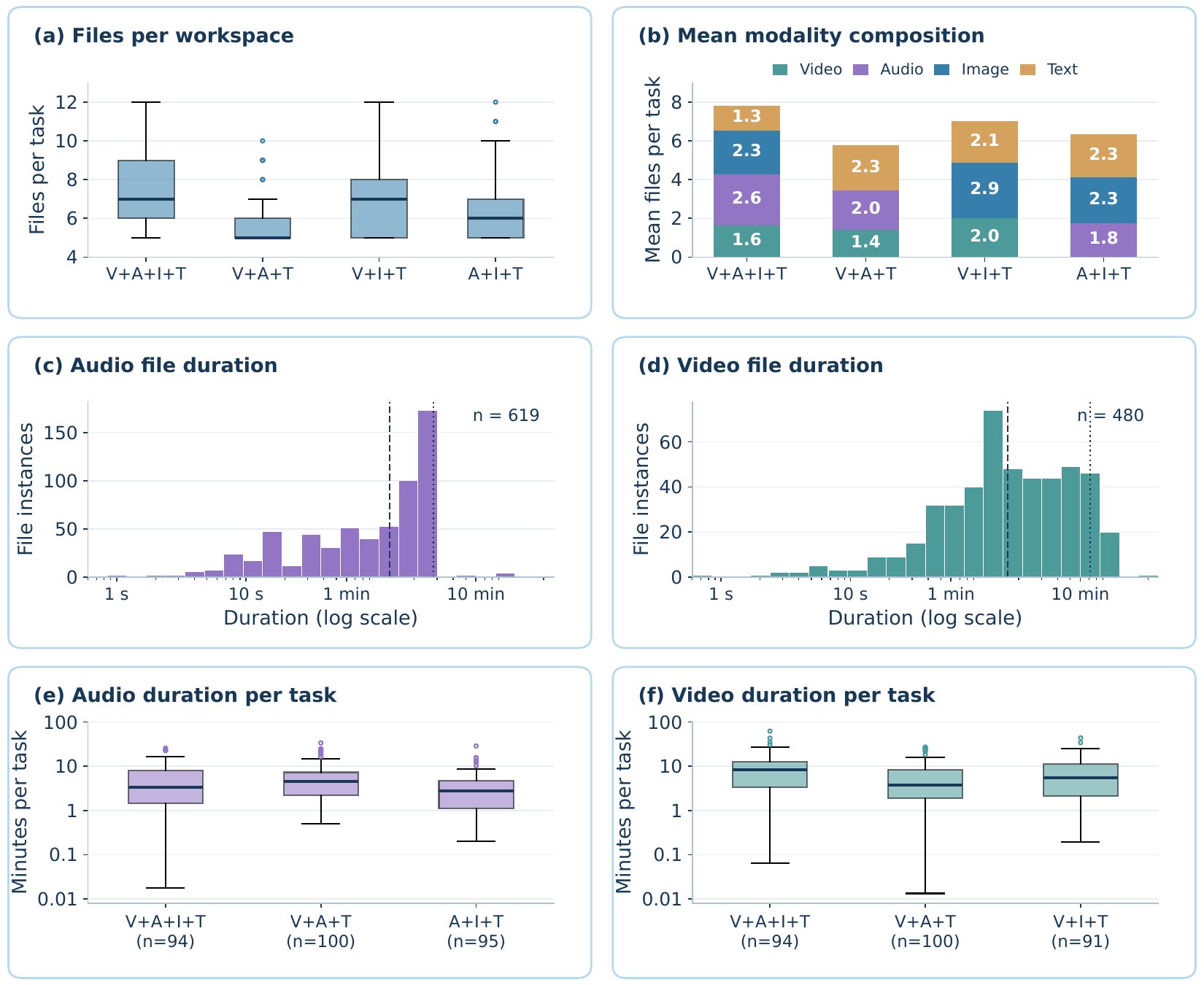}
    \caption{\textbf{Workspace structure and media distributions.}
    (a--b) File counts and mean modality composition by workspace type.
    (c--d) Audio and video file durations with shared logarithmic bins;
    dashed and dotted lines mark the median and 90th percentile.
    (e--f) Total media duration per task, including only tasks with that
    modality. Boxplots show medians, interquartile ranges (IQR), whiskers
    within 1.5 times the IQR of the box boundaries, and outliers.}
    \label{fig:app_b3_workspace_media}
\end{figure*}

\paragraph{Workspace structure.}
The 2,552 initial file instances comprise 480 videos, 619 audio files, 696 images, and 757 text files, including templates and reference materials.
Gold outputs, rubrics, and agent-generated files are excluded; reused files are counted separately across tasks. Mean file counts are 7.82, 5.76, 7.00, and 6.36 for V+A+I+T, V+A+T, V+I+T, and A+I+T, respectively (Figure~\ref{fig:app_b3_workspace_media}(a--b)). For delivery, 360 tasks require one file, 18 require two, and two require three, yielding 402 outputs.

\paragraph{Media durations.}
Median audio and video durations are 128.94 and 164.78 seconds; their 90th percentiles are 281.63 and 713.68 seconds. File instances total 24.63 audio hours and 37.38 video hours. Video audio tracks are not additionally counted as standalone audio. Per-task totals sum the corresponding initial media durations (Figure~\ref{fig:app_b3_workspace_media}(e--f)).

\begin{figure*}[!ht]
    \centering
    \includegraphics[width=\linewidth]{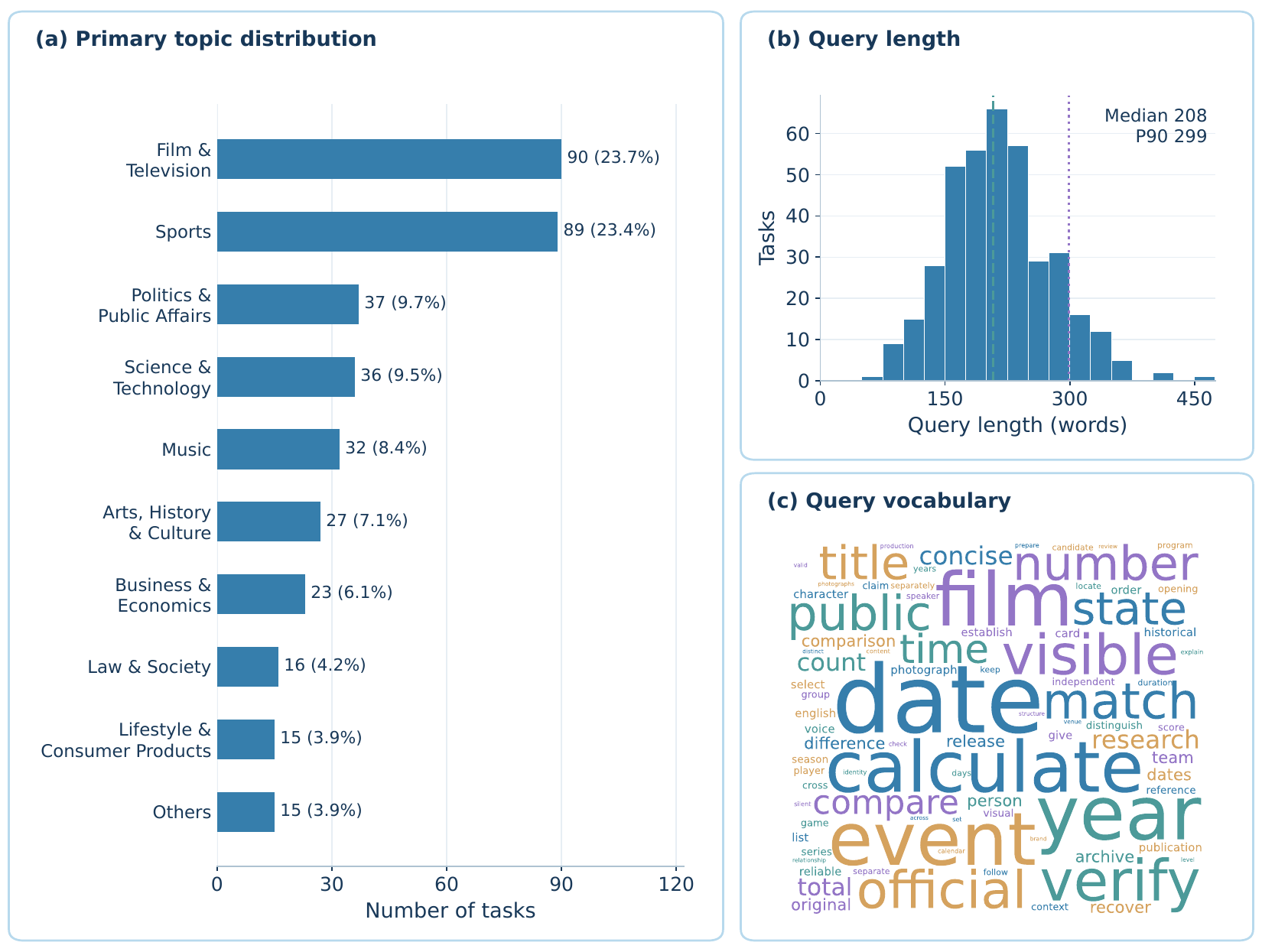}
    \caption{\textbf{Topic coverage and query characteristics.}
    (a) Primary topic counts and percentages across 380 tasks, with each task counted once. (b) Query length distribution in words; dashed and dotted lines indicate the median and 90th percentile, respectively. (c) A word cloud highlighting frequently occurring terms in the task instructions.}
    \label{fig:app_b3_topics_queries}
\end{figure*}

\paragraph{Primary topics.}
Tasks cover 10 primary topics (Figure~\ref{fig:app_b3_topics_queries}(a)). Labels are obtained through agent-based semantic annotation of queries, with supporting text or reference outputs consulted when needed. Each task receives one primary label according to its final research objective and main deliverable. Only primary labels are counted. Film \& Television and Sports contain 90 and 89 tasks, respectively, accounting for 47.1\% combined. The other 201 tasks span eight topics, including Politics \& Public Affairs, Science \& Technology, and Music.

\paragraph{Query length.}
Complete original instructions range from 74 to 454 words, with a mean
of 213.8, a median of 208, and a 90th percentile of 299
(Figure~\ref{fig:app_b3_topics_queries}(b)). Counts are based on
whitespace-separated units and retain filenames, paths, and delivery
requirements. 

\paragraph{Query vocabulary.}
Figure~\ref{fig:app_b3_topics_queries}(c) provides a lexical overview of the task instructions, highlighting recurring terms related to task content and requested operations. Common function words and file-related expressions are omitted to emphasize informative terms.

\subsection{Validating the Workspace Setting}
\label{app:workspace_task_dependencies}

\begin{table}[t]
\centering
\caption{Task requirements across 380 tasks. Strong cross-file dependency is a subset of cross-file integration. The properties may overlap.}
\label{tab:workspace_task_dependencies}
\vspace{4pt}
\setlength{\tabcolsep}{3.8pt}
\renewcommand{\arraystretch}{1.06}
\scriptsize
\begin{adjustbox}{max width=\linewidth}
\begin{tabular}{lrr}
\toprule
\textbf{Task property} & \textbf{Tasks} & \textbf{Proportion} \\
\midrule
Cross-file integration & 380 & 100.00\% \\
\quad Strong cross-file dependency & 297 & 78.16\% \\
Media-file processing (L2) & 169 & 44.47\% \\
Intermediate-artifact reuse & 103 & 27.11\% \\
\bottomrule
\end{tabular}
\end{adjustbox}
\end{table}

We examine the role of an accessible, operable file workspace in
task completion through a structural audit and a comparison
without workspace access. The audit characterizes dependencies
among files and operations on them; the comparison tests whether
initial multimodal context, web search, and general computation
can substitute for continued interaction with workspace files.

All 380 tasks require \emph{cross-file integration}: information
from multiple workspace files must be combined to produce the
requested result. We further characterize strong cross-file
dependency, media-file processing, and intermediate-artifact
reuse. Table~\ref{tab:workspace_task_dependencies} summarizes
their prevalence. 

\paragraph{Strong cross-file dependency.}
A strong cross-file dependency occurs when information from one
input file is needed to select, locate, decode, align, or interpret
evidence in another. This property appears in 297 tasks
(78.16\%). For example, an automotive research task requires
identifying a manufacturer in an audio recording and using its
name to select an auto-show photograph with a matching overhead
brand sign. 

\paragraph{Media-file processing.}
The 169 L2 tasks (44.47\%) require transformation or analysis of \emph{media files} to obtain information needed for the requested result. These files serve as inputs to tool or code operations. Such processing may directly return measurements or other analytical results without producing a new file. L2 therefore concerns required operations on media files, independently of whether a generated file is reused.

\paragraph{Intermediate-artifact reuse.}
Intermediate-artifact reuse occurs when a file generated during execution serves as input to a subsequent operation. This property appears in 103 tasks (27.11\%). For example, a cross-program research task requires using supplied calibration parameters to separate two audio channels with crosstalk and saving the recovered recording as an audio file. The agent then inspects this file to identify the speaker, program, and spoken content for subsequent verification. The generated file thus connects signal processing to semantic analysis. Unlike media-file processing alone, this property requires a generated file to be used in a later step.

\paragraph{Comparison without workspace access.}
We compare \emph{Workspace}, which allows continued inspection and processing of input files and reuse of intermediate files, with \emph{No workspace}, which removes file-based access to task materials. To make the initial materials available after removing the workspace, the latter supplies all initial file contents in the first model request: text is provided in full, and images, audio, and video as native multimodal inputs. Filenames identify the materials but do not provide access to file objects. The supplied inputs remain in context throughout execution.

Both settings permit multi-turn reasoning, web search,
general-purpose computation, and final output-file creation.
In \emph{No workspace}, computation can operate on information
available in context, but tools and code cannot read, transform,
or analyze workspace files or consume generated intermediate
files. Final output-file creation is retained to preserve the
same deliverable requirements and evaluation.

Both settings use Gemini 3 Flash Preview on all 380 tasks, with
at most 50 model requests per task, including the initial request.
Workspace scores are reused from the original evaluation.

\paragraph{Results and implications.}
As shown in Table~\ref{tab:workspace_first_turn_input},
Workspace scores 55.62 versus 33.43 for No workspace, a gap of
22.19 points. The observed gap is larger on L2 than L1 tasks
(25.06 versus 19.89 points), for tasks with strong cross-file
dependency than without it (23.29 versus 18.25 points), and for
tasks involving intermediate-artifact reuse than without it
(25.56 versus 20.94 points). These descriptive differences are
consistent with the corresponding file-processing and dependency
requirements.

Together, the structural audit and experimental comparison
support the substantive role of the workspace setting in this
benchmark. Under the tested model and budget, providing all
initial materials as context, alongside web search and general
computation, does not match performance with continued access
to files for inspection, processing, and intermediate-artifact
reuse.

\begin{table}[t]
\centering
\caption{Workspace versus No workspace for Gemini 3 Flash
Preview. No workspace supplies all initial file contents in the
first request and removes subsequent workspace-file access and
intermediate-file reuse, while retaining final output-file
creation. Scores are task averages on a 0--100 scale, and
$\Delta$ denotes the Workspace advantage in score points.
Difficulty, strong cross-file dependency, and intermediate-artifact
reuse each define a separate partition of the same 380 tasks.}
\label{tab:workspace_first_turn_input}

\setlength{\tabcolsep}{6pt}
\renewcommand{\arraystretch}{1.08}
\small

\begin{adjustbox}{max width=\linewidth}
\begin{tabular}{lrrrr}
\toprule
\textbf{Subset}
& \textbf{Tasks}
& \textbf{Workspace}
& \shortstack[r]{\textbf{No}\\\textbf{workspace}}
& \textbf{$\Delta$} \\
\midrule
All tasks
& 380 & 55.62 & 33.43 & +22.19 \\
\midrule
L1
& 211 & 60.87 & 40.98 & +19.89 \\
L2
& 169 & 49.06 & 24.00 & +25.06 \\
\midrule
Strong cross-file dependency
& 297 & 53.99 & 30.70 & +23.29 \\
No strong cross-file dependency
& 83 & 61.45 & 43.20 & +18.25 \\
\midrule
Intermediate-artifact reuse
& 103 & 44.55 & 18.99 & +25.56 \\
No intermediate-artifact reuse
& 277 & 59.74 & 38.80 & +20.94 \\
\bottomrule
\end{tabular}
\end{adjustbox}
\end{table}

\section{Tools and Evaluation Settings}
\label{app:eval-settings}

\begin{table*}[!htbp]
\centering
\caption{Overview of tool interfaces across model configurations. An asterisk (\textsuperscript{*}) marks optional parameters; a dash denotes no parameters.}
\label{tab:tool-specifications}
\begingroup
\footnotesize
\setlength{\tabcolsep}{3pt}
\renewcommand{\arraystretch}{1.12}
\newcommand{\toolname}[1]{\texttt{#1}}
\begin{tabularx}{\textwidth}{@{}p{0.105\textwidth}p{0.245\textwidth}X p{0.245\textwidth}@{}}
\toprule
\textbf{Category} & \textbf{Tool Name} & \textbf{Description} & \textbf{Parameters} \\
\midrule
\multirow[t]{4}{*}{Workspace}
& \toolname{list\_\allowbreak workspace\_\allowbreak files}
& List visible input files and files generated for the current task.
& --- \\
& \toolname{get\_metadata}
& Retrieve file format, size, duration, resolution, and aspect ratio, as applicable.
& \toolname{path} \\
& \toolname{read\_text\_file}
& Read a text file with an optional character limit.
& \toolname{path}, \toolname{max\_chars}\textsuperscript{*} \\
& \toolname{write\_text\_file}
& Write text to the current task's output directory.
& \toolname{path}, \toolname{content} \\
\midrule
\multirow[t]{5}{*}{Perception}
& \toolname{read\_image}
& Provide an image as visual input to the model.
& \toolname{path} \\
& \toolname{read\_video}
& Provide a video as multimodal input to the model.
& \toolname{path} \\
& \toolname{read\_audio}
& Provide audio to models supporting native audio input.
& \toolname{path} \\
& \toolname{transcribe\_audio}
& Transcribe speech from an audio file using Whisper.
& \toolname{path} \\
& \toolname{ocr\_image}
& Extract visible text from an image with confidence information.
& \toolname{path} \\
\midrule
\multirow[t]{12}{*}{Processing}
& \toolname{clip\_video}
& Extract a temporal video segment and return its file path.
& \toolname{path}, \toolname{start}, \toolname{end} \\
& \toolname{clip\_audio}
& Extract a temporal audio segment and return its file path.
& \toolname{path}, \toolname{start}, \toolname{end}, \toolname{sample\_rate}\textsuperscript{*} \\
& \toolname{extract\_audio\_\allowbreak from\_video}
& Extract the complete video audio track and return its file path.
& \toolname{path}, \toolname{sample\_rate}\textsuperscript{*} \\
& \toolname{rotate\_image}
& Create an image rotated by the specified angle.
& \toolname{path}, \toolname{angle} \\
& \toolname{flip\_image}
& Create a horizontally or vertically flipped image.
& \toolname{path}, \toolname{direction} \\
& \toolname{change\_audio\_speed}
& Change audio speed while preserving pitch.
& \toolname{path}, \toolname{speed}, \toolname{sample\_rate}\textsuperscript{*} \\
& \toolname{change\_audio\_pitch}
& Shift audio pitch while approximately preserving duration.
& \toolname{path}, \toolname{semitones}, \toolname{sample\_rate}\textsuperscript{*} \\
& \toolname{diarize\_speakers}
& Return timestamped speaker labels using pyannote Community-1.
& \toolname{path}, \toolname{start}\textsuperscript{*}, \toolname{end}\textsuperscript{*} \\
& \toolname{speaker\_match}
& Compare two clips using WeSpeaker embeddings and return similarity and a match decision.
& \toolname{reference\_path}, \toolname{candidate\_path},
  \toolname{reference\_start}\textsuperscript{*}, \toolname{reference\_end}\textsuperscript{*},
  \toolname{candidate\_start}\textsuperscript{*}, \toolname{candidate\_end}\textsuperscript{*} \\
& \toolname{python\_compute}
& Execute Python for numerical, date, and tabular calculations.
& \toolname{code}, \toolname{timeout}\textsuperscript{*} \\
& \toolname{python\_media\_analyzer}
& Execute Python to measure and analyze local media features.
& \toolname{code}, \toolname{timeout}\textsuperscript{*} \\
\midrule
Web Search
& \toolname{web\_search}
& Search the public Web and return titles, URLs, and snippets.
& \toolname{query}, \toolname{top\_k}\textsuperscript{*} \\
& \toolname{image\_query}
& Search images using text and return image results with source URLs for visual inspection.
& \toolname{query}, \toolname{num} \\
& \toolname{page\_browse}
& Retrieve webpage content from specified URLs for source verification.
& \toolname{url}, \toolname{summarizer}, \toolname{max\_chars\_per\_page} \\
\midrule
Control
& \toolname{finish}
& Signal task completion and provide a completion message.
& \toolname{message}, \toolname{summary}\textsuperscript{*} \\
\bottomrule
\end{tabularx}
\endgroup
\end{table*}

\subsection{Tool Specifications}
\label{app:tool-specifications}

The tool interfaces are organized into five functional categories: workspace access, multimodal perception, content processing, Web search, and execution control. Table~\ref{tab:tool-specifications} summarizes their functions and parameters. Workspace tools support file discovery and text I/O; perception tools provide media inputs or recognition results; processing tools support media transformations, speaker analysis, and programmatic computation. Search and control tools retrieve external evidence and terminate execution, respectively.

Media-reading tools provide files as model inputs without the descriptions, transcripts, or entity annotations produced during dataset construction.Clipping, transformation, and audio-extraction tools return paths to derived files, which the agent can subsequently inspect using reading or analysis tools. Native omni-modal agents receive audio directly, whereas VLM+ASR agents obtain speech transcriptions through Whisper~\citep{radford2023whisper}; video speech is transcribed after extracting its audio track. Speaker diarization uses the pyannote \texttt{speaker-diarization-community-1} pipeline~\citep{pyannoteCommunity1} to produce timestamped speaker labels. Speaker matching uses the pipeline's \texttt{wespeaker-voxceleb-resnet34-LM} embedding model~\citep{wang2023wespeaker} to encode reference and candidate clips and compare them using cosine similarity. 

\subsection{Model and Evaluation Settings}
\label{app:model-and-evaluation-settings}


\paragraph{Model access and inference settings.}
We access Gemini~\citep{google2026gemini35flash,google2025gemini3flash,google2025gemini25pro,gemini31pro2026}, MiMo~\citep{xiaomi2026mimov25}, Qwen3.5-Omni~\citep{qwen35omni2026},
Qwen3.5~\citep{qwen2026qwen35_122b,qwen2026qwen35_35b},
Qwen3.6~\citep{qwen2026apiplatform},
Qwen3.7~\citep{qwen37max2026},
Qwen3-VL~\citep{bai2025qwen3vl},
Kimi~\citep{moonshot2026kimik26},
and MiniMax~\citep{minimax2026m3} through hosted APIs.
MiniCPM-o-4.5~\citep{cui2026minicpmo45},
Nemotron-3-Nano-Omni~\citep{nvidia2026nemotronomni},
and Qwen3-Omni~\citep{qwen3omni2025}
are deployed locally using vLLM~\citep{kwon2023vllm}.
Thinking mode is enabled for all evaluated models except
MiMo-V2.5, Qwen3-VL-30B-A3B-Instruct, Qwen3-VL-8B-Instruct,
MiniCPM-o-4.5, Nemotron-3-Nano-Omni, and Qwen3-Omni.

\paragraph{Agent context and instructions.}
The initial user message contains the task instruction and the workspace path. The same workspace serves as the writable output directory. The agent discovers files and inspects their contents through tools during execution. The system prompt specifies how the agent should gather evidence, use tools, write deliverables, and terminate. Figure~\ref{fig:agent-system-prompt} presents the native-agent prompt template; the tool inventory is provided in Appendix~\ref{app:tool-specifications}. Audio-access instructions are adapted to the model interface: native omni-modal agents inspect audio directly, whereas VLM+ASR agents obtain speech transcriptions through the ASR tool. The same box includes the initial user-message template, with braced placeholders for task-specific values and repeated tool-use notes omitted.

\begin{figure*}[htbp]
\centering
\fbox{%
\begin{minipage}{\dimexpr\textwidth-2\fboxsep-2\fboxrule\relax}
\small
\setlength{\parindent}{0pt}
\setlength{\parskip}{0.45em}
\textbf{Agent system prompt}

You are an Omni-modal agent.

You complete user tasks in a workspace containing videos, audio, images,
and text assets.

The first user message gives only:
\begin{itemize}
\setlength{\itemsep}{0pt}
\item the user query
\item the workspace path
\item the writable output directory, which is the same workspace path
\end{itemize}
The first user message does not include metadata, media contents, or
text-file contents. Use list\_workspace\_files to discover available files
and the corresponding reading tools to inspect their contents.

\textbf{Rules}
\begin{enumerate}
\setlength{\itemsep}{0.2em}
\item Use list\_workspace\_files when you need to refresh which input, output, or media-
intermediate files are currently available. It scans only the workspace and does not inspect
file contents or arbitrary host directories.
\item Use read\_video/read\_audio/read\_image for workspace media evidence. These tools
attach raw media to the next model turn; inspect the attached media directly. Do not
invent file contents from file names or metadata.
\item Call read\_video at most once in each model turn. If several videos are required,
read one video now and request the others in later turns. This limit applies only to
read\_video. read\_audio and read\_image may be batched when useful. Use
clip\_video/clip\_audio first when the task identifies a narrow segment of a long
file.
\item Use read\_text\_file for workspace text assets. Do not assume text-file contents from
filenames or descriptions.
\item Use get\_metadata(path) on specific workspace file paths to resolve indirect
references such as duration, keywords, format, aspect ratio, or file role hints. It
does not list metadata for all files at once.
\item Use clip\_video or clip\_audio when the question points to a timestamped segment or
asks to locate a file by segment behavior. These tools return generated paths only;
call read\_video or read\_audio on the returned path before relying on the clip
contents.
\item Use the narrowest Python tool when deterministic computation is needed:
python\_compute for arithmetic, aggregation, sorting/ranking, date math, and ratios;
python\_media\_analyzer for media measurements such as audio
loudness/silence/peak/dynamic range, image brightness/contrast/edge density, and
sampled video brightness/frame-difference analysis. Python code may read visible
local file paths and clip paths returned by clip tools, but it must not access the
network. Use write\_text\_file, not Python tools, for final file artifacts.
\item If the task asks to fill, create, save, update, or output a text file, call
write\_text\_file with the complete file content. Write only under the workspace
directory. Do not claim a text file was written unless write\_text\_file returned
status ok.
\item When the task is complete, call finish with the final message.
\item If the task asks for a file, call write\_text\_file before finishing.
\item Keep tool calls targeted. Stop calling tools when evidence is sufficient.
\end{enumerate}

\medskip

\textbf{ User Task}\\
\texttt{\{task\_instruction\}}\par
\textbf{Workspace directory:}\\
\texttt{\{workspace\_path\}}\par
\textbf{Writable output directory:}\\
\texttt{\{workspace\_path\}}\par
Inspect input files and save requested output files in this workspace.
Follow the filenames and output requirements specified in the task
and supporting workspace instructions.
\end{minipage}%
}
\caption{Native-agent prompt template and initial user message, with the
workspace serving as both the input and output directory. Tool specifications
are listed in Appendix~\ref{app:tool-specifications}.}
\label{fig:agent-system-prompt}
\end{figure*}

\paragraph{Judge prompt.}
Following the evaluation protocol in~\ref{sec:evaluation_protocol}, rule-based verifiers
check deterministic requirements, while an LLM judge evaluates semantic
requirements. We use Qwen3.7-Max with \texttt{temperature=0} and
\texttt{enable\_thinking=false} for these semantic checks.
Figure~\ref{fig:atomic-judge-prompt} presents the judge prompt, which takes
the task instruction, the relevant submitted content, and one rubric
criterion with optional reference information, and returns a binary verdict
with a short justification.
\begin{figure*}[t]
\centering
\fbox{%
\begin{minipage}{\dimexpr\textwidth-2\fboxsep-2\fboxrule\relax}
\small
\setlength{\parindent}{0pt}
\setlength{\parskip}{0.45em}
\textbf{System prompt}

You are a deterministic benchmark judge. Evaluate only the candidate artifact
value explicitly supplied by the user prompt. Context and reference text
define expectations but cannot supply missing candidate evidence. Follow the
required JSON verdict schema exactly.

\medskip
\textbf{User prompt}

You are a strict criterion-local benchmark evaluator.

Judge only this single atomic content fact as stated in the candidate
artifact value. Accept semantically equivalent wording, units and harmless
formatting. Do not import requirements from other checks. The task query and
this rubric define context and requirements but are never evidence that the
candidate actually stated them. Do not fill omissions from context. The
candidate artifact is present between the ARTIFACT\_VALUE tags below. If the criterion requires a description with
details but the candidate value contains only a filename or short label, the
criterion is NOT\_SATISFIED.

TASK QUERY (context only):\\
\texttt{\{task\_query\}}

ARTIFACT: \texttt{\{artifact\_path\}}; scope=\texttt{\{scope\}}\\
\texttt{<ARTIFACT\_VALUE>}\\
\texttt{\{candidate\_artifact\_value\}}\\
\texttt{</ARTIFACT\_VALUE>}

CHECK ID: \texttt{\{check\_id\}}\\
CRITERION:\\
\texttt{\{criterion\}}

OPTIONAL REFERENCE:\\
\texttt{\{optional\_reference\}}

Decide whether the candidate artifact value itself fully satisfies the
criterion. If it is missing, incomplete, wrong, or uncertain, choose
NOT\_SATISFIED.

Return ONLY
\texttt{\{"verdict": "SATISFIED"|"NOT\_SATISFIED",}
\texttt{"reason": "<one short sentence consistent with the verdict>"\}}.
\end{minipage}%
}
\caption{Judge prompt for an atomic content check. Braced placeholders are
replaced with the corresponding task, artifact, and rubric values.}
\label{fig:atomic-judge-prompt}
\end{figure*}

\section{OmniFocus Implementation Details}
\label{app:omnifocus}

We implement OmniFocus by introducing two tools into the agent harness: \texttt{commit\_media\_evidence} preserves task-relevant media observations as textual records, while \texttt{update\_task\_state} maintains task requirements, acquired evidence, and completion progress. Together, they provide working memory for subsequent reasoning and action selection.

\subsection{Multimodal Working Memory}
\label{app:omnifocus-media}

When inspecting media returned by tools, the agent uses \texttt{commit\_media\_evidence} to record task-relevant observations and their sources, preferably alongside its next actions. After this response, the harness replaces the inspected raw media with textual placeholders, retaining the observations for subsequent interactions. Source files remain in the workspace and can be revisited when the records are insufficient for later decisions.

The tool takes four required arguments: \texttt{media\_action\_id} identifies the media from the current read, \texttt{source\_path} specifies its source, \texttt{observations} is a list of task-relevant observation strings, and \texttt{task\_relevance} explains their relationship to the user request.

\subsection{Explicit Task-State Maintenance}
\label{app:omnifocus-state}

As evidence or progress changes, the agent uses \texttt{update\_task\_state} to update a checklist of requirements, their status, findings, and remaining gaps. Each submission replaces the previous checklist in full while retaining completed items. The harness appends the latest checklist to every model request, making updates visible from the next request onward. The checklist reflects the model's assessment of progress, rather than verified completion, and does not directly determine termination.

The tool takes one required argument, \texttt{items}, an array of objects representing the complete checklist. Each item has three required fields: \texttt{requirement} describes the task requirement; \texttt{status} is \texttt{todo}, \texttt{done}, or \texttt{blocked}; and \texttt{note} preserves findings, source or artifact paths, and remaining gaps.

\subsection{Execution Procedure}
\label{app:omnifocus-execution}

Algorithm~\ref{alg:omnifocus-execution} summarizes how media records and task state are maintained across interaction turns.

\begin{algorithm}[t]
\caption{OmniFocus execution workflow}
\label{alg:omnifocus-execution}
\begin{algorithmic}[1]
\Require User task $Q$, workspace $W$, tool set $\mathcal{A}$, maximum rounds $T_{\max}$
\Ensure Available task artifacts $F$
\State Initialize context $C$ from the system prompt, $Q$, workspace path, and $\mathcal{A}$
\State $S \gets \varnothing$ \Comment{Initial task state}
\For{$t=1,\ldots,T_{\max}$}
    \State Form request $R$ from $C$ and the latest state $S$
    \If{$C$ contains pending raw media}
        \State Add guidance to $R$ to record observations alongside the next actions
    \EndIf
    \State $a \gets \Call{Agent}{R}$
    \State Process media-record submissions in $a$
    \If{$a$ contains a structurally valid task-state update}
        \State Replace $S$ with the submitted checklist
    \EndIf
    \If{raw media was inspected in this turn}
        \State Replace it in $C$ with placeholders; retain source files in $W$
    \EndIf
    \State Execute task actions in $a$ through the harness, obtaining observations $o$
    \State Append $a$ and tool responses to $C$, retaining submitted textual records
    \State Leave new media in $o$ for inspection on the next request
    \If{the harness accepts task completion}
        \State \textbf{break}
    \EndIf
\EndFor
\State \Return $\Call{CollectArtifacts}{W,Q}$
\end{algorithmic}
\end{algorithm}

\section{Supplementary Results and Analysis}
\label{app:supplementary_results}

\subsection{Detailed Benchmark Results}
\label{app:detailed_results}

Table~\ref{tab:omni_workspace_main_results_components} supplements the main results with separate Content and Format scores for each workspace type, difficulty level, and the full benchmark. Overall Format scores exceed Content scores for all evaluated agents. The breakdown also reveals differences between agents with similar composite scores. For example, Kimi K2.6 achieves a higher Content score than Gemini 3 Flash Preview (53.25 vs.\ 51.71), but a lower Format score (72.34 vs.\ 90.80), resulting in a slightly lower composite score (55.16 vs.\ 55.62). Reporting both components therefore distinguishes content fulfillment from format compliance, which may be obscured by the composite score alone.

\begin{table*}[t]
    \centering
    \caption{
    \textbf{Detailed results on Omni-Workspace.}
    We report Content (C), Format (F), and composite scores
    ($S=0.9C+0.1F$) on a 0--100 scale, averaged over tasks
    in each subset.
    Overall weights all 380 tasks equally.
    Bold indicates the highest score in each column.
    }
    \label{tab:omni_workspace_main_results_components}

    \setlength{\tabcolsep}{3pt}
    \renewcommand{\arraystretch}{1.08}
    \resizebox{\textwidth}{!}{%
    \begin{tabular}{@{}l*{4}{rrr!{\hspace{4pt}}}*{2}{rrr!{\hspace{4pt}}}rrr@{}}
        \toprule
        \multirow{3}{*}{\textbf{Model}}
        & \multicolumn{12}{c}{\textbf{Workspace Type}}
        & \multicolumn{6}{c}{\textbf{Difficulty Levels}}
        & \multicolumn{3}{c}{\textbf{Overall}} \\
        \cmidrule(lr){2-13} \cmidrule(lr){14-19} \cmidrule(lr){20-22}
        & \multicolumn{3}{c}{V+A+I+T}
        & \multicolumn{3}{c}{V+A+T}
        & \multicolumn{3}{c}{V+I+T}
        & \multicolumn{3}{c}{A+I+T}
        & \multicolumn{3}{c}{L1}
        & \multicolumn{3}{c}{L2}
        & \multicolumn{3}{c}{All 380 tasks} \\
        \cmidrule(lr){2-4} \cmidrule(lr){5-7}
        \cmidrule(lr){8-10} \cmidrule(lr){11-13}
        \cmidrule(lr){14-16} \cmidrule(lr){17-19} \cmidrule(lr){20-22}
        & C & F & S & C & F & S & C & F & S & C & F & S
        & C & F & S & C & F & S & C & F & S \\

        \midrule
        \rowcolor{gray!12}
        \multicolumn{22}{c}{\textit{Native Omni-Modal Agents}} \\
        \midrule

        Gemini 3.5 Flash
        & \textbf{65.85} & 90.16 & \textbf{68.29} & 59.00 & 86.24 & 61.72 & 63.41 & 84.47 & 65.52 & \textbf{66.09} & 89.32 & \textbf{68.42} & \textbf{70.95} & 89.17 & \textbf{72.77} & 54.26 & 85.54 & 57.38 & \textbf{63.53} & 87.56 & \textbf{65.93} \\

        Gemini 3.1 Pro Preview
        & 61.63 & \textbf{95.46} & 65.01 & 48.07 & \textbf{91.70} & 52.43 & 55.45 & 86.80 & 58.59 & 60.69 & 93.16 & 63.93 & 62.11 & \textbf{92.32} & 65.13 & 49.15 & \textbf{91.20} & 53.36 & 56.35 & \textbf{91.82} & 59.89 \\

        Gemini 3 Flash Preview 
        & 55.54 & 94.11 & 59.40 & 47.84 & 85.94 & 51.65 & 51.36 & \textbf{91.17} & 55.34 & 52.34 & 92.28 & 56.33 & 57.49 & 91.34 & 60.87 & 44.50 & 90.13 & 49.06 & 51.71 & 90.80 & 55.62 \\

        Gemini 2.5 Pro
        & 47.79 & 93.44 & 52.36 & 40.27 & 82.84 & 44.53 & 42.21 & 90.76 & 47.06 & 45.25 & 89.22 & 49.65 & 49.38 & 90.56 & 53.50 & 36.92 & 86.94 & 41.93 & 43.84 & 88.95 & 48.35 \\

        MiMo-V2.5
        & 48.59 & 86.63 & 52.40 & 39.25 & 74.93 & 42.82 & 38.78 & 72.07 & 42.11 & 45.28 & 77.64 & 48.51 & 49.60 & 83.84 & 53.02 & 34.66 & 70.30 & 38.23 & 42.96 & 77.82 & 46.44 \\

        Qwen3.5 Omni Plus
        & 38.64 & 81.26 & 42.90 & 31.22 & 71.74 & 35.27 & 36.57 & 69.55 & 39.87 & 36.14 & 72.30 & 39.76 & 40.64 & 79.58 & 44.53 & 29.24 & 66.38 & 32.95 & 35.57 & 73.71 & 39.38 \\

        Qwen3.5 Omni Flash
        & 21.78 & 71.07 & 26.71 & 9.28 & 65.68 & 14.92 & 25.00 & 64.76 & 28.98 & 30.88 & 62.14 & 34.01 & 26.05 & 71.49 & 30.60 & 15.89 & 58.94 & 20.20 & 21.53 & 65.91 & 25.97 \\

        MiniCPM-o-4.5
        & 9.53 & 61.60 & 14.74 & 9.95 & 60.06 & 14.96 & 5.83 & 60.31 & 11.28 & 8.54 & 52.71 & 12.96 & 10.03 & 63.97 & 15.43 & 6.60 & 52.03 & 11.14 & 8.51 & 58.66 & 13.52 \\

        Nemotron-3-Nano-Omni
        & 5.11 & 36.15 & 8.22 & 6.77 & 36.17 & 9.71 & 2.93 & 28.57 & 5.49 & 4.76 & 41.33 & 8.41 & 6.37 & 39.07 & 9.64 & 3.15 & 31.35 & 5.97 & 4.94 & 35.63 & 8.01 \\

        Qwen3-Omni-Instruct
        & 4.45 & 12.80 & 5.29 & 3.73 & 15.98 & 4.96 & 4.59 & 24.65 & 6.60 & 4.29 & 18.71 & 5.73 & 4.79 & 17.41 & 6.05 & 3.59 & 18.63 & 5.09 & 4.26 & 17.95 & 5.63 \\

        \midrule
        \rowcolor{gray!12}
        \multicolumn{22}{c}{\textit{VLM-Based Agents with ASR}} \\
        \midrule

        Qwen3.7-Plus
        & 64.07 & 94.13 & 67.08 & \textbf{59.69} & 87.35 & \textbf{62.46} & \textbf{63.52} & 87.11 & \textbf{65.88} & 60.61 & \textbf{93.46} & 63.90 & 66.21 & 91.10 & 68.70 & \textbf{56.56} & 89.74 & \textbf{59.88} & 61.92 & 90.50 & 64.78 \\
        
        Qwen3.6-Plus
        & 60.05 & 87.37 & 62.78 & 53.76 & 81.15 & 56.50 & 57.32 & 83.06 & 59.90 & 51.28 & 82.25 & 54.37 & 60.39 & 83.99 & 62.75 & 49.50 & 82.71 & 52.82 & 55.55 & 83.42 & 58.34 \\

        Kimi K2.6
        & 57.16 & 78.60 & 59.31 & 52.98 & 76.38 & 55.32 & 60.43 & 80.25 & 62.41 & 42.77 & 54.30 & 43.92 & 59.31 & 78.49 & 61.23 & 45.67 & 64.65 & 47.57 & 53.25 & 72.34 & 55.16 \\

        MiniMax-M3
        & 48.91 & 85.58 & 52.58 & 50.88 & 81.82 & 53.98 & 48.60 & 87.67 & 52.51 & 45.18 & 87.11 & 49.37 & 52.50 & 87.21 & 55.97 & 43.34 & 83.31 & 47.33 & 48.42 & 85.47 & 52.13 \\

        Qwen3.5-122B-A10B
        & 48.98 & 80.20 & 52.10 & 42.96 & 71.75 & 45.84 & 50.09 & 77.41 & 52.82 & 51.31 & 85.26 & 54.70 & 53.34 & 81.63 & 56.17 & 41.87 & 74.76 & 45.16 & 48.24 & 78.57 & 51.27 \\

        Qwen3.5-35B-A3B
        & 38.05 & 72.55 & 41.50 & 41.04 & 64.24 & 43.36 & 38.99 & 73.32 & 42.42 & 38.42 & 77.11 & 42.29 & 44.89 & 74.36 & 47.83 & 32.00 & 68.35 & 35.63 & 39.15 & 71.69 & 42.41 \\

        Qwen3-VL-30B-A3B-Instruct
        & 17.14 & 55.95 & 21.02 & 18.34 & 55.11 & 22.02 & 13.22 & 49.85 & 16.89 & 16.85 & 57.01 & 20.87 & 18.16 & 59.05 & 22.25 & 14.30 & 48.89 & 17.76 & 16.45 & 54.53 & 20.25 \\

        Qwen3-VL-8B-Instruct
        & 11.88 & 65.50 & 17.24 & 8.74 & 50.36 & 12.90 & 9.60 & 50.17 & 13.66 & 13.14 & 62.80 & 18.11 & 12.55 & 61.64 & 17.46 & 8.67 & 51.59 & 12.96 & 10.82 & 57.17 & 15.46 \\

        \bottomrule
    \end{tabular}%
    }
\end{table*}

\subsection{Audio-Focused Subsets and Evaluation Settings}
\label{app:audio_analysis_details}

\paragraph{Subset definitions and statistics.}
We examine two audio-focused subsets according to the information required for task completion. The Non-Speech Information (NSI) subset contains 42 tasks that require acoustic evidence not captured by the words in a speech transcript. Relevant evidence includes timbral characteristics, environmental sounds, and musical cues, which agents must use to distinguish sources, interpret events, or identify task-relevant media. Here, ``non-speech information'' refers to the information needed, rather than requiring the recording itself to contain no speech: a recording may mix spoken content with acoustic cues that are necessary for the requested result.

The Multimodal Speaker Understanding (MSU) subset contains 49 tasks that require establishing who speaks across segments or recordings and attributing relevant content to the correct speaker. Agents must distinguish speakers, recognize when the same speaker reappears, and connect the corresponding utterances with visual or textual evidence as required by the task. Recovering the spoken words alone is insufficient when the result depends on which person made a statement or whether voices in different recordings belong to the same speaker. For both subsets, inclusion requires that the respective information be necessary for task completion; the mere presence of audio or multiple voices is insufficient.

\paragraph{Evaluation configurations.}
We distinguish three configurations according to how agents obtain audio information. Native omni-modal agents inspect audio directly through \texttt{read\_audio}, allowing them to jointly interpret spoken content, voice characteristics, and non-speech sounds from the audio input. VLM+ASR agents obtain spoken content as text through \texttt{transcribe\_audio} and use \texttt{diarize\_speakers} and \texttt{speaker\_match} for speaker analysis. Diarization identifies speaker turns within a recording, while speaker matching compares voices across audio segments to support speaker correspondence. These tools provide information that helps agents associate transcribed content with the relevant speakers and connect it with visual or textual evidence. Tool implementations are described in Appendix~\ref{app:tool-specifications}.

The ASR-only configuration restricts audio perception to speech transcriptions, without access to \texttt{diarize\_speakers} or \texttt{speaker\_match}. Transcriptions provide spoken content, so agents must reason from the transcribed words and the other available task evidence. Python-based audio analysis (via \texttt{python\_media\_analyzer}) is also prohibited, preventing agents from using code to extract additional acoustic features or compare speaker voices. This configuration examines task performance when audio evidence is available only through its transcribed speech content.

\paragraph{Detailed subset results.}
Table~\ref{tab:audio-subset-detailed-results} reveals different effects of tool-mediated audio perception on NSI and MSU. On MSU, speaker-analysis tools provide speaker boundaries and cross-segment voice correspondence that complement speech transcripts. These signals help VLM-based agents associate spoken content with the correct speakers, narrowing the gap to native audio perception. Qwen3.7-Plus reaches 64.67, compared with 66.77 for Gemini 3.5 Flash, while Qwen3.6-Plus nearly matches Gemini 3.1 Pro (63.09 vs.\ 63.14). Under ASR-only, their scores fall to 53.78 and 49.71, decreases of 10.89 and 13.38 points respectively, consistent with their usefulness for speaker attribution.

On NSI, a performance gap remains between the evaluated native omni-modal agents and the tool-augmented VLM-based agents: the two native models score 62.39–63.02, compared with 47.75–52.89 for the VLM-based agents with ASR and additional audio-analysis capabilities. This contrast highlights the stronger performance of the evaluated native omni-modal agents on tasks requiring acoustic evidence beyond transcribed speech.


\begin{table}[htbp]
\centering
\small
\definecolor{AudioAblationBg}{HTML}{F2F5F7}
\definecolor{AudioDeltaText}{HTML}{B33A3A}
\setlength{\tabcolsep}{10pt}
\renewcommand{\arraystretch}{1.15}
\caption{Performance on audio-focused subsets. Bold marks the highest score in each subset. Parenthesized values show score changes relative to the corresponding VLM+ASR configuration.}
\label{tab:audio-subset-detailed-results}
\begin{tabular}{@{}lcc@{}}
\toprule
\textbf{Model / configuration} & \textbf{NSI} $\uparrow$
& \textbf{MSU} $\uparrow$ \\
\midrule
Gemini 3.5 Flash (native) & 62.39 & \textbf{66.77} \\
Gemini 3.1 Pro (native) & \textbf{63.02} & 63.14 \\
\midrule
Qwen3.7-Plus (ASR + tools) & 52.89 & 64.67 \\
\rowcolor{AudioAblationBg}
Qwen3.7-Plus (ASR-only) & 51.37 & 53.78 \\
& {\tiny\bfseries\boldmath\color{AudioDeltaText} $(-1.52)$}
& {\tiny\bfseries\boldmath\color{AudioDeltaText} $(-10.89)$} \\
\addlinespace[4pt]
Qwen3.6-Plus (ASR + tools) & 47.75 & 63.09 \\
\rowcolor{AudioAblationBg}
Qwen3.6-Plus (ASR-only) & 43.32 & 49.71 \\
& {\tiny\bfseries\boldmath\color{AudioDeltaText} $(-4.43)$}
& {\tiny\bfseries\boldmath\color{AudioDeltaText} $(-13.38)$} \\
\bottomrule
\end{tabular}
\end{table}


\subsection{Error Taxonomy and Annotation Protocol}
\label{app:error_analysis_details}

\paragraph{Error taxonomy.}
We distinguish six error categories in two groups. Local execution errors concern individual operations in multimodal perception, search, and tool invocation. Task-state drift errors concern failures to maintain requirements, acquired evidence, and execution progress across steps. Table~\ref{tab:error-taxonomy} defines the six categories and gives illustrative examples.

\begin{table}[htbp]
\centering
\small
\caption{Error categories, definitions, and illustrative examples.}
\label{tab:error-taxonomy}
\renewcommand{\arraystretch}{1.15}
\setlength{\tabcolsep}{4pt}
\begin{tabularx}{\linewidth}{@{}>{\raggedright\arraybackslash}p{0.19\linewidth}>{\raggedright\arraybackslash}X>{\raggedright\arraybackslash}p{0.25\linewidth}@{}}
\toprule
\textbf{Category} & \textbf{Definition} & \textbf{Example} \\
\midrule
\multicolumn{3}{@{}l}{\textit{Local execution errors}} \\
\addlinespace[2pt]
Multimodal perception & Incorrectly recognizing or transcribing task-relevant content in images, video, or audio. & Mistaking the person who speaks in a video for another visible person. \\
Search & Failing to find required information, or selecting a result about the wrong entity, source, time, or version. & Selecting the release date of a different film with the same title. \\
Tool invocation & Using incorrect parameters, paths, code, or operations that prevent a tool from performing the intended action correctly. & Processing media using an incorrect file path. \\
\midrule
\multicolumn{3}{@{}l}{\textit{Task-state drift errors}} \\
\addlinespace[2pt]
Requirement omission & Omitting or replacing an explicit requirement, or incorrectly treating it as completed. & Analyzing only one recording when the task requires comparing two. \\
Evidence loss or misuse & Losing, overwriting, misattributing, or incorrectly using previously acquired information during subsequent execution. & Swapping two people's dates in the final artifact despite having obtained the correct associations. \\
Redundant tool calls & Making at least five consecutive calls, including the first, to the same tool for the same entity and the same or a closely related goal, when prior results or feedback make the repetition unnecessary. & Repeatedly requesting an already obtained fact with identical parameters and no further verification need. \\
\bottomrule
\end{tabularx}
\end{table}

The boundary between search and evidence loss or misuse depends on when the failure occurs: search covers finding or selecting evidence, whereas evidence loss or misuse concerns information the agent has already acquired. Similarly, rechecking to resolve a genuine uncertainty is not redundant; the redundant-tool-call category requires repeated invocations without substantive progress. This includes continued calls after sufficient information has been obtained and repeated failed calls without a meaningful change in approach.

\paragraph{Annotation.}
We analyze 760 baseline trajectories, comprising one trajectory for each of the same 380 tasks run by Gemini 3 Flash Preview and one for each run by Gemini 3.1 Pro Preview. We use Qwen3.7-Max as the annotation model to identify errors from the task instruction, model messages, tool calls and responses, and final artifacts; reference answers are used for verification. Category counts represent error instances rather than trajectories, and a trajectory may contain multiple errors.


\subsection{Further Analysis of OmniFocus}
\label{app:omnifocus_analysis_details}

\paragraph{Efficiency metric definitions.}

Input and output tokens are accumulated over all model requests within a task; output tokens include thinking tokens reported in the usage records, and $k$ denotes 1,000 tokens. Tool Calls exclude internal media-record and task-state updates. Their token consumption, cost, and elapsed time remain included. Cost is estimated from input and output token usage using the corresponding model's token prices, without accounting for caching or other pricing adjustments. It covers agent-model inference only, excluding judging and other tool costs. Time denotes per-task elapsed time.

\paragraph{Component-level effects.}
Removing task-state maintenance decreases the score from 60.73 to 57.10 for Gemini 3 Flash Preview and from 64.95 to 63.23 for Gemini 3.1 Pro Preview. It also increases mean tool calls by 6.55 and 1.89, respectively, and turns by 2.11 and 1.29. The time effect is less uniform: full OmniFocus is 118.01 seconds faster for Flash, whereas the variant without task state is 13.20 seconds faster for Pro. These comparisons associate task-state maintenance with higher scores and fewer interactions.

Removing memory management has a smaller reported score effect, reducing scores by 0.68 and 0.75 points relative to full OmniFocus. Its larger effect is on context and cost: mean input tokens rise from 249.20k to 426.53k for Flash and from 271.07k to 580.11k for Pro, while estimated cost rises from \$0.1467 to \$0.2323 and from \$0.7358 to \$1.3826. The corresponding execution times rise from 218.38 to 550.94 seconds and from 564.08 to 821.65 seconds. Tool calls and turns do not change uniformly across the two models, consistent with memory management primarily limiting context growth rather than directly reducing the number of actions.

\paragraph{Relation to error categories.}
We annotate baseline and OmniFocus trajectories for Gemini 3 Flash Preview and Gemini 3.1 Pro Preview on the same 380 tasks per model, using the six categories defined in Appendix~\ref{app:error_analysis_details}. Figure~\ref{fig:error-ablation} shows a larger reduction in the summed counts of task-state drift errors than in local execution errors (22.4\% versus 7.9\%). Requirement omission accounts for the largest absolute decrease among the six categories (272 to 191), whereas multimodal perception changes little (414 to 407). This pattern is consistent with OmniFocus helping agents maintain task requirements across steps, while multimodal perception errors remain largely unchanged.

\begin{figure}[htbp]
\centering
\includegraphics[width=\linewidth]{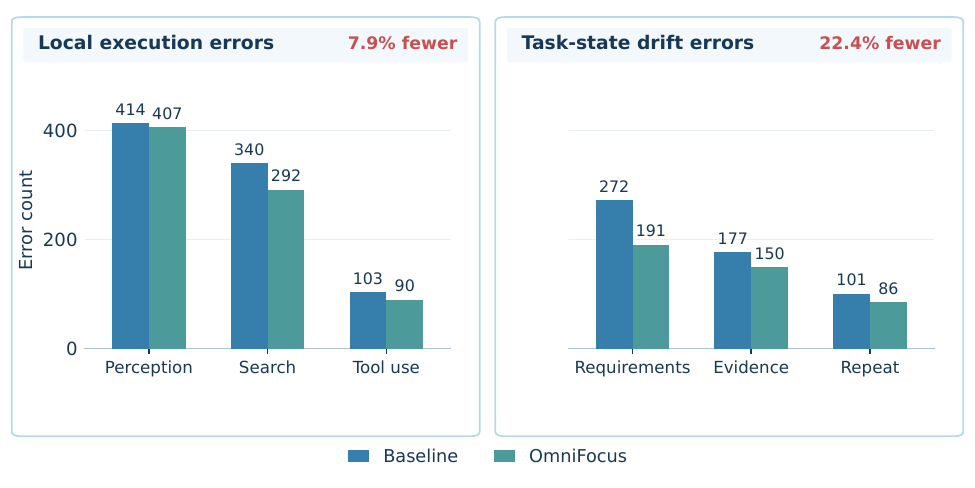}
\caption{\textbf{Error counts before and after OmniFocus.}
Percentages compare the sums of the three category counts in each group.}
\label{fig:error-ablation}
\end{figure}

\subsection{Cross-Judge Agreement and Ranking Robustness}
\label{app:cross_judge_consistency}

\paragraph{Experimental setup.}
To examine sensitivity to the choice of semantic judge, we use Gemini 3.5 Flash and Claude Opus 4.8~\citep{anthropic2026claudeopus48} to re-score results and compare their judgments with the original Qwen3.7-Max judgments. We hold task instructions, rubrics, extracted candidate content, and reference information fixed, changing only the judge for semantic checks. Programmatic checks and score aggregation remain unchanged.

\paragraph{Agreement on atomic checks.}
As shown in Table~\ref{tab:cross_judge_consistency}, pairwise agreement ranges from 93.23\% to 95.49\%, with Cohen's $\kappa$ between 0.865 and 0.910. Both positive and negative agreement exceed 92\%, indicating consistency for satisfied as well as unsatisfied criteria. Qwen3.7-Max and Claude Opus 4.8 exhibit the highest agreement among the three pairs.

\paragraph{Ranking robustness.}
Model rankings are highly correlated across judges: Spearman's $\rho$ ranges from 0.9897 to 0.9959, and Kendall's $\tau_b$ from 0.9477 to 0.9739. All three judges agree on the ordering of the top three models, while the remaining differences involve local rank changes. These results support the robustness of relative model rankings on the compared task set.  The high agreement in atomic judgments and model rankings across three judge families provides empirical support for the robustness of our rubric-based evaluation protocol to the choice of semantic judge.

\begin{table}[htbp]
\centering
\small
\caption{Cross-judge agreement and model ranking correlations.
Qwen, Gemini, and Claude denote Qwen3.7-Max, Gemini 3.5 Flash,
and Claude Opus 4.8, respectively.}
\label{tab:cross_judge_consistency}
\setlength{\tabcolsep}{5pt}
\renewcommand{\arraystretch}{1.1}
\resizebox{\linewidth}{!}{%
\begin{tabular}{@{}lrrr@{}}
\toprule
Metric & Qwen--Gemini & Qwen--Claude & Gemini--Claude \\
\midrule
Atomic agreement & 93.23\% & 95.49\% & 93.39\% \\
Cohen's $\kappa$ & 0.865 & 0.910 & 0.868 \\
Positive agreement & 93.47\% & 95.43\% & 93.67\% \\
Negative agreement & 92.97\% & 95.55\% & 93.09\% \\
Spearman's $\rho$ & 0.9897 & 0.9897 & 0.9959 \\
Kendall's $\tau_b$ & 0.9477 & 0.9477 & 0.9739 \\
\bottomrule
\end{tabular}%
}
\end{table}

\section{Case Studies}
\label{app:case_studies}

\subsection{Case Study I: BookTube Publication Tracing}
\label{app:case_study_booktube}

\begin{figure*}[!ht]
    \centering
    \includegraphics[width=\linewidth,height=0.9\textheight,keepaspectratio]
        {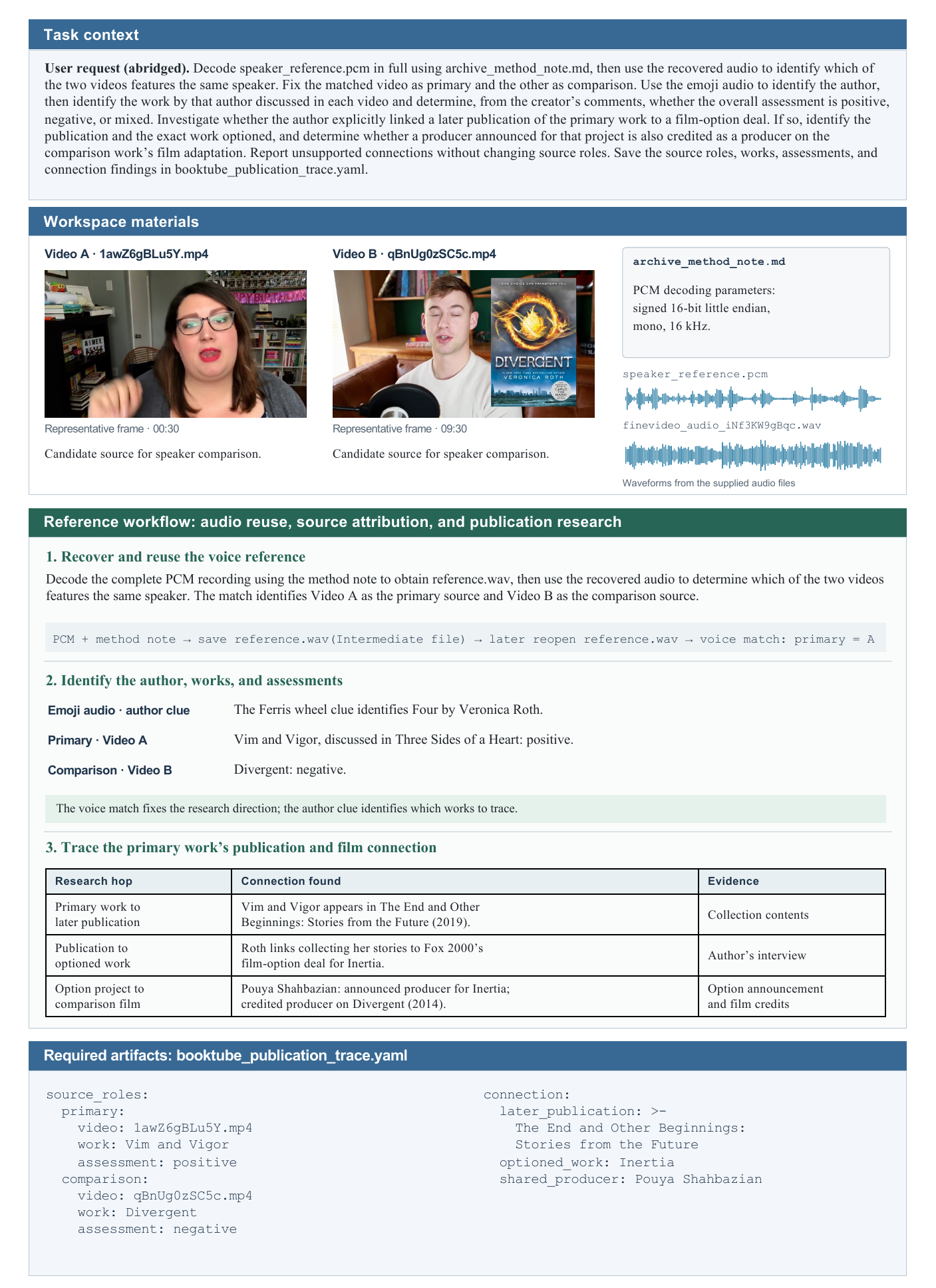}
    \caption{\textbf{BookTube publication tracing.} Task materials and
    reference workflow linking PCM decoding and audio reuse, cross-modal
    source attribution, and verification of publication, film-option,
    and producer connections, with findings recorded in a structured
    YAML report.}
    \label{fig:case_study_booktube}
\end{figure*}

Figure~\ref{fig:case_study_booktube} illustrates a task that requires
decoding a PCM recording using a method note and reusing the resulting
audio to establish primary and comparison video sources through speaker
matching. The agent then uses an emoji-audio clue to identify the shared
author and determines the work discussed and the creator's overall
assessment in each video. It investigates the author's stated link
between the primary work's later publication and a film-option deal,
and checks whether a producer announced for that project is also
credited on the comparison work's film adaptation. The findings are
recorded in a YAML report. This case highlights how media processing,
intermediate-artifact reuse, and cross-file evidence guide subsequent
Web research and structured artifact creation.

\subsection{Case Study II: Cross-Program Provenance Audit}
\label{app:case_study_cross_program}

\begin{figure*}[!ht]
    \centering
    \includegraphics[width=\linewidth,height=0.9\textheight,keepaspectratio]
        {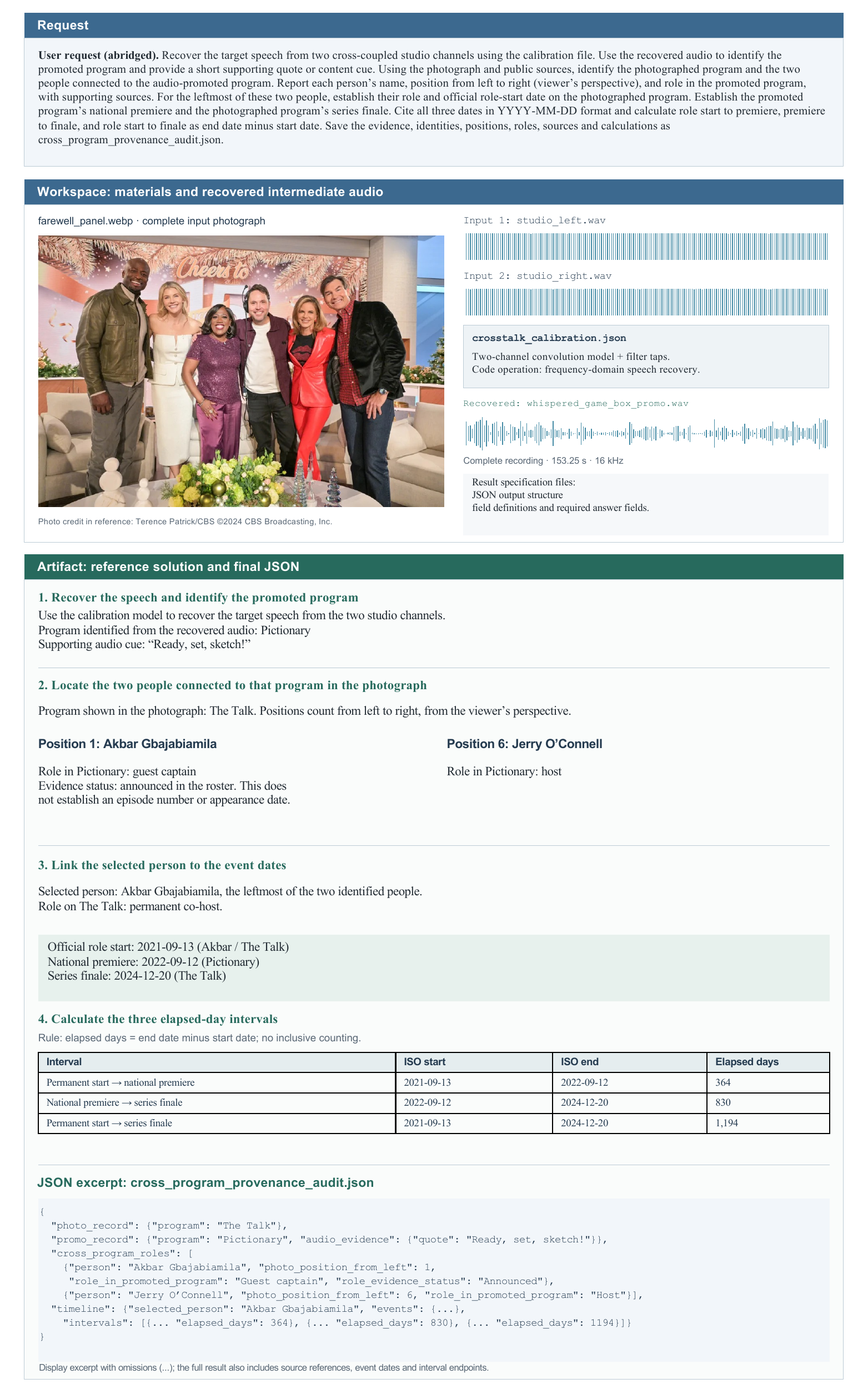}
    \caption{\textbf{Cross-program provenance audit.}
    Task request, input materials, recovered intermediate audio, and
    reference solution linking a \emph{Pictionary} promotion to two
    people in a photograph from \emph{The Talk}. The figure summarizes
    their cross-program roles, three event dates and the elapsed-day
    intervals between them, and an excerpt of the JSON report.}
    \label{fig:case_study_cross_program}
\end{figure*}

Figure~\ref{fig:case_study_cross_program} presents a cross-program
provenance audit that begins with calibrated speech recovery from two
cross-coupled audio channels and preservation of the complete recording.
The recovered audio provides a cue for identifying the promoted program,
which is then linked to a photographed program through publicly sourced
identities, viewer-relative positions, and personnel roles. In the
reference solution, \emph{Pictionary} is connected to \emph{The Talk}
through Akbar Gbajabiamila and Jerry O'Connell, whose roles in
\emph{Pictionary} are recorded as announced guest captain and host,
respectively. For Gbajabiamila, the leftmost of the two, the timeline
connects his official start as a permanent co-host of \emph{The Talk},
\emph{Pictionary}'s national premiere, and \emph{The Talk}'s series finale.
The three elapsed-day intervals are computed as end date minus start
date and recorded with the supporting evidence and sources in a JSON
report. This case highlights reuse of recovered audio as an intermediate
artifact and cross-modal evidence reconciliation under explicit
positional, role, and temporal constraints.

\clearpage

\end{document}